\documentclass{article} 
\usepackage{iclr2027_conference,times}

\usepackage{amsmath,amsfonts,bm}

\def\eqref#1{equation~\ref{#1}}

\def\1{\bm{1}}

\DeclareMathAlphabet{\mathsfit}{\encodingdefault}{\sfdefault}{m}{sl}
\SetMathAlphabet{\mathsfit}{bold}{\encodingdefault}{\sfdefault}{bx}{n}

\usepackage{hyperref}
\usepackage{booktabs}
\usepackage{graphicx}
\usepackage{url}
\usepackage{multirow}
\usepackage{enumitem}

\usepackage{xcolor}
\usepackage{soul}
\sethlcolor{yellow}


\hypersetup{hidelinks}
\iclrfinalcopy

    \title{LENS: Replacing LayerNorm and Softmax in Transformers with Learnable Normalization States}

\author{Yuhao Liu, Salim Ullah \& Akash Kumar \\
Chair of Embedded Systems, Ruhr-University Bochum, Germany\\
\texttt{\{yuhao.liu,salim.ullah,akash.kumar\}@ruhr-uni-bochum.de}
}

\begin{document}

\maketitle
\pagestyle{plain}
\thispagestyle{plain}

\begin{abstract}
    \emph{LayerNorm} and attention~\emph{Softmax} are critical operations in Transformers. However, they cannot be efficiently mapped to~\emph{General Matrix Multiplication} (GEMM) computations on hardware. Therefore, modern hardware accelerator designs, such as NPUs on ASIC/FPGA, consume a large amount of hardware resources or chip area to implement their specific arithmetic units. Building on previous work about~\emph{LayerNorm},~\emph{Softmax}, and their alternatives, we introduce~\emph{Learnable Normalization States} (LENS), a framework that reformulates~\emph{LayerNorm} as tiny MLPs by abstracting a common structure from related methods and extends this method to~\emph{Softmax}, which moves more of the computation in~\emph{Transformers} to GEMM as a hardware-friendly architecture. We evaluate LENS on image classification with~\emph{Vision Transformers} (ViT) and autoregressive language modeling with~\emph{Pythia} models, including generalization tests on~\emph{WikiText} and~\emph{LAMBADA}. Across these tasks, LENS matches or exceeds baseline performance in several vision settings and retains most of the baseline performance in the remaining vision and language experiments. These findings may suggest that hand-designed normalization and attention functions are not the only choice as alternatives to~\emph{LayerNorm} and~\emph{Softmax}. LENS also provides a basis for exploring Transformer architectures for resource-limited accelerators. Source code is available at \href{https://github.com/liuyh-Horizon/LENS}{Github}
\end{abstract}

\section{Introduction}
\label{introduction}
\emph{Transformer}~\citep{transformer} has been widely adopted in recent research as a foundational network architecture across various scenarios. Following the rapid expansion of model sizes in~\emph {Transformer}-based networks for vision and language tasks, massive compute requirements have driven and accelerated the development of hardware accelerator designs. Considering that~\emph {General Matrix Multiplication} (GEMM) is the most important and intensive computation workload during network inference, modern neural network hardware accelerators, whether \emph{Graphics Processing Units} (GPUs) or \emph{Neural Processing Units} (NPUs), usually consume a massive amount of hardware resources or chip area to implement the matrix computation core, such as the \emph{Tensor Cores} on \emph{NVIDIA} GPUs~\citep{nvidia_blackwell} and \emph{Systolic Arrays} on NPUs. 

However, in the~\emph{Transformer} architecture, \emph{Layer Normalization} (\emph{LayerNorm})~\citep{layernorm} and \emph{Softmax} in the attention layers are two heterogeneous nonlinear operators that cannot be directly mapped to GfEMM on hardware. As a result, hardware accelerators for~\emph{Transformer} have to implement the additional special arithmetic units for \emph{LayerNorm} and \emph{Softmax} or map them onto the existing functional cores. For instance, \emph{NVIDIA} GPUs accelerate the reciprocal, square root, and exponential operations in \emph{LayerNorm} and \emph{Softmax} on \emph{Special Function Units} (SFUs) and execute the rest of the standard arithmetic operations on CUDA cores~\citep{nvidia_fermi}. NPUs also need to customize the specific \emph{LayerNorm} and \emph{Softmax} units to further accelerate related computations on \emph{Application-Specific Integrated Circuits} (ASICs) or \emph{Field-Programmable Gate Arrays} (FPGAs).

\begin{table}[t]
    \centering
    \caption{Hardware Resource Consumption of Systolic Array, LayerNorm, and Softmax Reported by~\citet{llm_fpga}}
    \vspace{5pt}
    \label{fpga_costs}
    \resizebox{0.90\columnwidth}{!}
    {
        \begin{tabular}{ccccccc}
            \toprule
                \multirow{2}{*}{\textbf{Kernel}}    & \multicolumn{3}{c}{\textbf{Hardware Resource Consumption}}    & \multicolumn{3}{c}{\textbf{Relative to Systolic Array}}   \\
                                                    & LUTs          & FFs           & DSPs                          & LUTs          & FFs           & DSPs                      \\ \midrule
                Systolic Array with DSP Packing     & 244439        & 79969         & 128                           & $100.00\%$    & $100.00\%$    & $100.00\%$                \\
                Softmax                             & 7447          & 4835          & 38                            & $3.05\%$      & $6.05\%$      & $29.69\%$                 \\
                LayerNorm                           & 12746         & 18751         & 80                            & $5.21\%$      & $23.45\%$     & $62.50\%$                 \\ \midrule
                Softmax + 2$\times$LayerNorm        & 32939         & 42337         & 198                           & $13.48\%$     & $52.94\%$     & $154.69\%$                \\
            \bottomrule
    \end{tabular}%
    }
    \vspace{-10pt}
\end{table}

As a result, designing a hardware accelerator for \emph{Transformers} involves a critical trade-off: allocating more hardware resources or chip area to implementing GEMM computation cores can speed up linear-layer inference but may leave \emph{LayerNorm}, \emph{Softmax}, and other heterogeneous nonlinear operators as a performance bottleneck. Allocating more hardware resources or chip area to heterogeneous nonlinear operators can reduce their inference latency. However, it may also reduce the available area for implementing matrix-computation cores and slow linear-layer inference. For instance,~\citet{llm_fpga} successfully implemented the FPGA-based \emph{Large Language Model} (LLM) accelerator for \emph{Bidirectional Encoder Representations from Transformers} (BERT)~\citep{bert} and \emph{Generative Pre-trained Transformer 2} (GPT-2)~\citep{gpt2} on an \emph{AMD Xilinx Alveo U280} FPGA device. They report that the \emph{LayerNorm} and \emph{Softmax} units do not significantly increase the overall latency in their fully pipelined design. However, as shown in~\autoref{fpga_costs}, the hardware resources consumed by two \emph{LayerNorm} units and one \emph{Softmax} unit in their accelerator are equivalent to $13.48\%$ of the LUTs, $52.94\%$ of the FFs, and $154.69\%$ of the DSPs used by a single systolic-array kernel. Therefore, given that the vast majority of computation in \emph{Transformers} is linear and can be mapped to GEMMs, this example suggests that hardware accelerators may still need to allocate a massive amount of hardware resources to prevent the \emph{LayerNorm} and \emph{Softmax} from becoming a performance bottleneck and increasing the inference latency, even though they account for only a small fraction of the total computation in \emph{Transformer} inference.

Previous work, such as \emph{Root Mean Square Layer Normalization} (RMSNorm)~\citep{rmsnorm}, \emph{Dynamic Tanh} (DyT)~\citep{dyt}, \emph{Dynamic erf} (Derf)~\citep{derf}, \emph{SeeDNorm}~\citep{seednorm}, \emph{SimA}~\citep{sima}, and \emph{I-ViT}~\citep{ivit}, has widely explored the different variants, alternatives, and approximations to simplify the \emph{LayerNorm} and \emph{Softmax}. Considering that most of the above-mentioned works match or exceed the performance of the native \emph{Transformer} architecture in different vision and language tasks, we therefore rethink the related works and have the following questions:

\begin{itemize}[leftmargin=2.5em]
    \item If those different replacements of \emph{LayerNorm} and \emph{Softmax} work properly in \emph{Transformers}, is there any shared underlying mechanism across some of these different methods?

    \item If there is a shared underlying mechanism across these methods, can we follow it and let training determine the appropriate functions to replace \emph{LayerNorm} and \emph{Softmax}, rather than relying on hand-designed mathematical formulas?

    \item Is it possible to reformulate the heterogeneous nonlinear operators of \emph{LayerNorm} and \emph{Softmax} to map more computations to GEMMs? 
\end{itemize}

Following these questions, we present the \emph{Learnable Normalization States} (LENS) in this work to replace both \emph{LayerNorm} and \emph{Softmax} with tiny \emph{Multilayer Perceptrons} (MLPs). The following are our contributions in this work:

\begin{itemize}[leftmargin=2.5em]
    \item We first outlined the mathematical formulas for \emph{LayerNorm} and its alternatives and discovered that they can be unified into a common functional form $\mathbf{a}(\mathbf{x})\odot\mathbf{x}+\mathbf{b}(\mathbf{x})$. We therefore adopted a tiny MLP with a two-neuron output layer to replace it. And then, we extended a similar mechanism to the \emph{Softmax} and also replaced it with two tiny MLPs.

    \item We tested the LENS architecture across various vision and language tasks using different datasets. The results show that LENS recovers most of the baseline performance and even exceeds it in several vision settings after retraining.
    
    \item We further conducted a preliminary analysis of the LENS training process, as well as the weight changes and function plots observed before and after replacing \emph{LayerNorm} and \emph{Softmax}. The results show that LENS is not merely a simple approximation of \emph{LayerNorm} and \emph{Softmax}. Instead, it underwent joint adaptation with the model weights during retraining, significantly changing some weight values and thereby learning a different functional mapping.
\end{itemize}

The goal of this work is to explore a hardware-design-friendly, GEMM-centric alternative to \emph{LayerNorm} and \emph{Softmax} that can shift more nonlinear computation in \emph{Transformers} onto GEMMs on accelerators. For ASIC design, considering that the design and manufacturing costs of network inference accelerator chips are always directly proportional to the chip area, replacing the \emph{LayerNorm} and \emph{Softmax} with MLPs to shift computation to the matrix-multiplication core and eliminating the corresponding hardware module to reduce chip area can yield a realistic benefit. For FPGA design, eliminating the corresponding hardware modules of \emph{LayerNorm} and \emph{Softmax} can also reduce the hardware resources and power consumption. Furthermore, our work gives a proof of concept suggesting that hand-designed mathematical functions are not the only choices for replacing \emph{LayerNorm} and \emph{Softmax}

\section{Mechanism Analysis and LENS}
\label{mechanism}
\subsection{Related Work}
\label{related_work}

To explore the potential shared mechanism, we summarized some representative works first, starting with methods that simplify or alter the \emph{LayerNorm}: 

\begin{itemize}[leftmargin=2.5em]
    \item \emph{RMSNorm}~\citep{rmsnorm} is a classic design in related works that removes the mean-centering operation in \emph{LayerNorm} and normalizes the input using the \emph{Root Mean Square} (RMS).
    \item \citet{dyt} present the \emph{Dynamic Tanh} (DyT) to replace \emph{LayerNorm} with an element-wise $\tanh(\alpha \mathbf{x}) \odot \gamma+\beta$, where $\alpha$ is a learnable scale factor, $\gamma$ and $\beta$ are learnable affine parameters. Therefore, DyT has no input-statistics computation, such as mean and variance, in \emph{LayerNorm}.
    \item \emph{Dynamic erf} (Derf)~\citep{derf} extends DyT by replacing \emph{tanh} with \emph{erf} and adding one additional learnable bias factor $s$. The function of Derf becomes $\operatorname{erf}(\alpha \mathbf{x}+s) \odot \gamma+\beta$.
    \item \emph{SeeDNorm}~\citep{seednorm} extends the \emph{RMSNorm}, adding a dynamically adjusted scaling coefficient based on the current input to the $\gamma$ in \emph{RMSNorm}.
\end{itemize}

And the following are some representative alternatives to \emph{Softmax}:

\begin{itemize}[leftmargin=2.5em]
    \item \emph{SimA}~\citep{sima} replaces \emph{Softmax} by applying $\ell_1$ normalization to each feature dimension of the query and key matrices across tokens as $\widehat{\mathbf{Q}}$ and $\widehat{\mathbf{K}}$. The attention output is then directly computed as $\widehat{\mathbf{Q}}\widehat{\mathbf{K}}^{\mathsf T}\mathbf{V}$.
    \item \citet{relu_attn} present the \emph{ReLU Attention}, which replaces \emph{Softmax} with an element-wise \emph{ReLU} function applied to $\mathbf{Q}\mathbf{K}^{\mathsf T}/\sqrt{d}$ and divides the result by the sequence length $N$. The complete formula is $\operatorname{ReLU}(\mathbf{Q}\mathbf{K}^{\mathsf T}/\sqrt{d})/N$. 
    \item \citet{soft} present the \emph{SOFT} to replace scaled dot-product \emph{Softmax} attention with an unnormalized \emph{Gaussian} kernel similarity function. It further applies the \emph{Nyström} low-rank decomposition and approximates the required \emph{Moore-Penrose} inverse via \emph{Newton-Raphson} iterations, achieving linear complexity with respect to the sequence length.
\end{itemize}

Furthermore, \emph{I-ViT}~\citep{ivit} proposes integer-only approximations to \emph{Softmax} and \emph{LayerNorm} for quantized \emph{Transformers}. It rewrites the exponential function in \emph{Softmax} from $e^x$ to $2^{x\log_2e}$ and adopts $1+\frac{1}{2}-\frac{1}{16}$ to approximate the $\log_2e$, which can be converted into additions and bitwise shifting. And for the \emph{LayerNorm}, it computes the mean and variance using integer arithmetic and adopts a fixed 10-step integer Newton iteration to approximate the square root. \emph{I-ViT} does not replace the functions of \emph{Softmax} and \emph{LayerNorm}, but approximates them for integer-only \emph{Transformer} inference. In addition, \mbox{\citet{dn}} proposes \emph{Dynamic Normalization} (DN), which learns the grouping and aggregation of normalization statistics. However, its online normalization still requires explicit computations of the mean and standard deviation and input-dependent division. Therefore, its complete operation cannot be directly mapped to GEMMs. \mbox{\citet{SAL-ViT}} presents \emph{SAL-ViT}, which introduces learnable scaling and shifting parameters into a quadratic \emph{Softmax} approximation for efficient private inference while retaining input-dependent row-sum normalization and division, which are also difficult to shift to the matrix-multiplication core.

\subsection{Converting LayerNorm to LENSNorm}

Summarizing the representative \emph{LayerNorm} alternatives listed in Section~\ref{related_work}, we find that: \emph{RMSNorm} and \emph{SeeDNorm} remove the variance in~\emph{LayerNorm} but still require computing the reciprocal, mean, and square root. DyT and Derf replace the reciprocal, mean, variance, and square root with \emph{tanh} or \emph{erf}, both of which are nonlinear functions. Furthermore, none of these four methods can be simply mapped to GEMMs either.

\begin{table}[t]
    \centering

    \caption{Unified formulation of \textit{LayerNorm} and its alternatives as $\mathbf{a}(\mathbf{x})\odot\mathbf{x}+\mathbf{b}(\mathbf{x})$ in \textit{LENSNorm}.} 
    \vspace{5pt}
    \label{formular_compare}
    \resizebox{1\columnwidth}{!}
    {
        \begin{tabular}{cccc}
            \toprule
            
            Methods
            & Original
            & $\mathbf{a}(\mathbf{x})$
            & $\mathbf{b}(\mathbf{x})$ \\ \midrule
            \addlinespace[2.5pt]
    
            \footnotesize{$\operatorname{LayerNorm}(\mathbf{x})$}
            & \footnotesize{$\frac{\mathbf{x}-\operatorname{Mean}(\mathbf{x})}{\sqrt{\operatorname{Var}(\mathbf{x})+\epsilon}} \odot \boldsymbol{\gamma}+\boldsymbol{\beta}$}
            & \footnotesize{$\frac{\boldsymbol{\gamma}}{\sqrt{\operatorname{Var}(\mathbf{x})+\epsilon}}$}
            & \footnotesize{$\frac{\sqrt{\operatorname{Var}(\mathbf{x})+\epsilon}\,\boldsymbol{\beta} - \operatorname{Mean}(\mathbf{x})\,\boldsymbol{\gamma}}{ \sqrt{\operatorname{Var}(\mathbf{x})+\epsilon} }$} \\
            \addlinespace[2.5pt]
            \cmidrule(lr){2-4}
    
            \addlinespace[2.5pt]
            \footnotesize{$\operatorname{RMSNorm}(\mathbf{x})$}
            & \footnotesize{$\frac{\mathbf{x}}{\sqrt{\operatorname{Mean}(\mathbf{x}\odot\mathbf{x})+\epsilon}} \odot \boldsymbol{\gamma}$}
            & \footnotesize{$\frac{\boldsymbol{\gamma}}{\sqrt{\operatorname{Mean}(\mathbf{x}\odot\mathbf{x})+\epsilon}}$}
            & \footnotesize{$0$} \\
            \addlinespace[2.5pt]
            \cmidrule(lr){2-4}
    
            \addlinespace[2.5pt]
            \footnotesize{$\operatorname{DyT}(\mathbf{x})$}
            & \footnotesize{$\operatorname{tanh}(\alpha\mathbf{x})\odot\boldsymbol{\gamma}+\boldsymbol{\beta}$}
            & \footnotesize{$\alpha\boldsymbol{\gamma}-\frac{\alpha^3}{3}\boldsymbol{\gamma}\odot\mathbf{x}^{\odot2}$}
            & \footnotesize{$\boldsymbol{\beta}$} \\
            \addlinespace[2.5pt]
            \cmidrule(lr){2-4}
    
            \addlinespace[2.5pt]
            \footnotesize{$\operatorname{Derf}(\mathbf{x})$}
            & \footnotesize{$\operatorname{erf}(\alpha\mathbf{x}+s)\odot\boldsymbol{\gamma}+\boldsymbol{\beta}$}
            & \footnotesize{$\mathcal{R}(\alpha\mathbf{x}+s)\odot\alpha\boldsymbol{\gamma}$}
            & \footnotesize{$\mathcal{R}(\alpha\mathbf{x}+s)\odot s\boldsymbol{\gamma}+\boldsymbol{\beta}$} \\
            \addlinespace[2.5pt]
            \cmidrule(lr){2-4}
    
            \addlinespace[2.5pt]
            \footnotesize{$\operatorname{SeeDNorm}(\mathbf{x})$}
            & \footnotesize{$\frac{\mathbf{x}}{\sqrt{\operatorname{Mean}(\mathbf{x}\odot\mathbf{x})+\epsilon}} \odot \left [ \sigma \left ( \mathbf{x}\boldsymbol{\beta}^{\mathsf T}  \right )\boldsymbol{\alpha}+\boldsymbol{\gamma} \right ]$}
            & \footnotesize{$\frac{\sigma \left ( \mathbf{x}\boldsymbol{\beta}^{\mathsf T}  \right )\boldsymbol{\alpha}+\boldsymbol{\gamma}}{\sqrt{\operatorname{Mean}(\mathbf{x}\odot\mathbf{x})+\epsilon}}$}
            & \footnotesize{$0$} \\
            \addlinespace[2.5pt]
            \midrule
            
            \addlinespace[2.5pt]
            \footnotesize{$\operatorname{LENSNorm}(\mathbf{x})$}
            & \footnotesize{$a(\mathbf{x})\mathbf{x}+b(\mathbf{x})$}
            & \footnotesize{$a(\mathbf{x})$}
            & \footnotesize{$b(\mathbf{x})$} \\
            \addlinespace[2.5pt]
            \bottomrule
    \vspace{-25pt}
    \end{tabular}%
    }
\end{table}

Therefore, we rethink the formula of \emph{LayerNorm}, $\frac{\mathbf{x}-\operatorname{Mean}(\mathbf{x})}{\sqrt{\operatorname{Var}(\mathbf{x})+\epsilon}} \odot\boldsymbol{\gamma}+\boldsymbol{\beta}$, and find that we can reorganize it in the format of $\mathbf{a}(\mathbf{x})\odot\mathbf{x}+\mathbf{b}(\mathbf{x})$, where $\mathbf{a}(\mathbf{x})$ denotes $\frac{\boldsymbol{\gamma}}{\sqrt{\operatorname{Var}(\mathbf{x})+\epsilon}}$ and $\mathbf{b}(\mathbf{x})$ denotes $\frac{\sqrt{\operatorname{Var}(\mathbf{x})+\epsilon}\,\boldsymbol{\beta} - \operatorname{Mean}(\mathbf{x})\,\boldsymbol{\gamma}}{ \sqrt{\operatorname{Var}(\mathbf{x})+\epsilon} }$. Following the same methods, as shown in~\autoref{formular_compare}, we can also convert the formulas of \emph{RMSNorm} and \emph{SeeDNorm} into the format of $\mathbf{a}(\mathbf{x})\odot\mathbf{x}+\mathbf{b}(\mathbf{x})$. Moreover, if we approximate \emph{tanh} as $\tanh(\mathbf{x})\approx\mathbf{x}-\frac{\mathbf{x}^3}{3}$ based on third-order \emph{Taylor Expansion} around zero and rationally approximate \emph{erf} as $\operatorname{erf}(\mathbf{x})\approx\operatorname{sign}(\mathbf{x})\left( 1-\frac{1}{(1+a_1|\mathbf{x}|+a_2|\mathbf{x}|^2+a_3|\mathbf{x}|^3+a_4|\mathbf{x}|^4)^4} \right)$, where $a_1=0.278393$, $a_2=0.230389$, $a_3=0.000972$, and $a_4=0.078108$~\citep{abramowitz1964handbook}, DyT and Derf can also be unified to the format of $\mathbf{a}(\mathbf{x})\odot\mathbf{x}+\mathbf{b}(\mathbf{x})$. The $\mathcal{R}(z)$ in~\autoref{formular_compare} denotes $\mathcal{R}(z)=\frac{1-(1+a_1|z|+a_2|z|^2+a_3|z|^3+a_4|z|^4)^{-4}}{|z|}$. The detailed mathematical derivation process can be found in Appendix~\ref{math_lens}.

Based on the above-mentioned mathematical conversion, we therefore unified \emph{LayerNorm} and its representative alternatives in the common format of $\mathbf{a}(\mathbf{x})\odot\mathbf{x}+\mathbf{b}(\mathbf{x})$. Inspired by this common format, we propose \emph{LENSNorm}, which uses two scalar normalization states $a(\mathbf{x})$ and $b(\mathbf{x})$ for input-dependent scaling and bias, in contrast to the vector-valued functions $\mathbf{a}(\mathbf{x})$ and $\mathbf{b}(\mathbf{x})$ above. 

\textbf{We parameterize these two scalar states with a tiny MLP as~\autoref{math_lensnorm} consisting of two \emph{Fully Connected} (FC) linear layers.} To limit additional parameters and computational overhead and avoid unnecessary over-parameterization, the first FC layer in LENSNorm has 32 neurons, followed by a \emph{ReLU} activation. The second FC layer has only two neurons that generate the scalar normalization states $a(\mathbf{x})$ and $b(\mathbf{x})$. To maintain numerical stability, $a(\mathbf{x})$ is constrained to $[0.05,12]$ using clamping, while $b(\mathbf{x})$ remains unconstrained. The two states are then broadcast across all hidden dimensions of the input token.

\vspace{-10pt}
\begin{equation}
    \small
    \begin{bmatrix}
        a(\mathbf{x}) \\
        b(\mathbf{x})
    \end{bmatrix}
    =\operatorname{MLP}_{\mathrm{N}}(\mathbf{x}),\qquad
    \operatorname{LENSNorm}(\mathbf{x})_i=\left[\operatorname{MLP}_{\mathrm{N}}(\mathbf{x})\right]_1 x_i+\left[\operatorname{MLP}_{\mathrm{N}}(\mathbf{x})\right]_2
    \label{math_lensnorm}
\end{equation}

We have also compared the designs of broadcasting two scale normalization states across all hidden dimensions versus training an individual pair of scale normalization states for each dimension. We define the methods of training individual scale normalization states for each dimension as \emph{LENSNorm+}. As shown in the~\mbox{\autoref{lensnorm_plus_results}} of Appendix~\mbox{\ref{lensnorm_plus_section}}, \emph{LENSNorm+} exceeds the accuracy of \emph{LENSNorm} in all experiments of CIFAR-10/100. However, the accuracy improvements of \emph{LENSNorm+} are only around $0.23\sim0.80$ percentage points. If the number of hidden dimensions is 128, \emph{LENSNorm} has 2 neurons in the last layer, whereas \emph{LENSNorm+} has 256 neurons. As a result, the number of weights in \emph{LENSNorm+} is around $2.95\times$ that in \emph{LENSNorm}. Therefore, given the tiny improvement in accuracy, the significant increase in \emph{LENSNorm+} weights may not be a sufficient trade-off. Considering the balance between accuracy, number of parameters, and computation efficiency on hardware, we adopted \emph{LENSNorm} in this work. 

\subsection{Converting Softmax to LENSMax}

Summarizing the representative \emph{Softmax} alternatives listed in Section~\ref{related_work}, we find that: \emph{SimA} applies $\ell_1$ normalization to each feature dimension of the query and key matrices before the score computation of $\mathbf{Q}\mathbf{K}^{\mathsf T}$. \emph {ReLU Attention} applies the element-wise \emph{ReLU} activation after the computation of $\mathbf{Q}\mathbf{K}^{\mathsf T}/\sqrt{d}$. \emph{SOFT} directly replaces the \emph{Softmax} with an unnormalized \emph{Gaussian} kernel and applies \emph{Newton–Raphson} iterations to approximate the required matrix inverse. Furthermore, computing the unnormalized \emph{Gaussian} kernel still involves exponential functions. As with the alternatives to \emph{LayerNorm}, most alternatives to \emph{Softmax} cannot be directly mapped to GEMMs. Considering that existing Softmax alternatives modify different stages of attention computation and do not exhibit a clear shared computational structure, instead of forcing these methods into a unified formulation, we extend the state-extraction and element-transformation principle of \emph{LENSNorm} to \emph{Softmax}.

Given an input vector $\mathbf{x}=[x_1,\ldots,x_N]$ as one row in $\mathbf{Q}\mathbf{K}^{\mathsf T}/\sqrt{d}$, the $i$-th output of \emph{Softmax} is $\operatorname{Softmax}(\mathbf{x})_i=\frac{\exp(x_i)}{\sum_{j=1}^{N}\exp(x_j)}$. If we define two functions as $\lambda(\mathbf{x})=\sum_{j=1}^{N}\exp(x_j)$ and $\mu(\mathbf{x})=\log\lambda(\mathbf{x})$, $\operatorname{Softmax}(\mathbf{x})_i=\exp\left(x_i-\mu(\mathbf{x})\right)=g\left(x_i,\mu(\mathbf{x})\right)$. Inspired by the $g\left(x_i,\mu(\mathbf{x})\right)$ format of \emph{Softmax}, we define \emph{LENSMax} as one row-state function $r(\mathbf{x})$ and one element-mapping function $e(x_i,r(\mathbf{x}))$. We then adopt two tiny MLPs to parameterize these two functions in \autoref{math_lensmax}. The detailed mathematical derivation process can be found in Appendix~\ref{math_max}.
\vspace{-5pt}

\begin{equation}
    \small
    \operatorname{LENSMax}(\mathbf{x})_i=\frac{\operatorname{MLP}_2\left(x_i,\operatorname{MLP}_1(\mathbf{x})\right)}{2^{\left\lfloor\log_2N\right\rfloor}}
    \label{math_lensmax}
\end{equation}

\textbf{The first MLP has two FC layers:} The first FC layer maps one row in $\mathbf{Q}\mathbf{K}^{\mathsf T}/\sqrt{d}$ to 32 neurons, followed by a \emph{ReLU} activation function. We retain the standard scaling factor $\sqrt{d}$, where $d$ denotes the head dimension. Since $1/\sqrt{d}$ is a fixed constant determined by the model architecture, it does not require an input-dependent division and can be implemented using constant scaling on hardware. The second FC layer has only one neuron to generate the row state. 

\textbf{The second MLP also has two FC layers:} The first FC layer maps one score and its corresponding row state from the first MLP to 16 neurons. The second FC layer has only one neuron to compute the new element result, followed by a \emph{Sigmoid} function. Then, the \emph{Sigmoid} result is divided by $2^{\left\lfloor\log_2N\right\rfloor}$, where $N$ denotes the number of valid elements in the row. Therefore, the sum of each \emph{LENSMax} output row is lower than 2. We use $2^{\left\lfloor\log_2N\right\rfloor}$, rather than $N$, also for hardware efficiency, which can also be easily converted to bitwise shifting in fixed-point format or to an exponent adjustment in floating-point format.

Compared with the previous representative alternatives to \emph{Softmax}, \emph{LENSMax} is hardware-friendly and also moves more computations to GEMMs. Although \emph{ReLU Attention} has a simpler computational structure than \emph{LENSMax}, its authors report no clear best choice among several point-wise functions, including \emph{ReLU}, \emph{ReLU$^2$}, \emph{GELU}~\citep{gelu}, and \emph{Softplus}, when combined with sequence-length scaling. They select \emph{ReLU} mainly because of its higher computational speed, while leaving open why the $L^{-1}$ scaling improves performance and whether a better activation function exists. These observations motivate us to learn the row-state and element mappings instead of manually selecting a fixed function.







\section{Experiments}
\label{experiments}
To analyze the network performance with \emph{LENSNorm} and \emph{LENSMax} after replacing \emph{LayerNorm} and \emph{Softmax}, we evaluated them in vision classification and causal language tasks. All training experiments were conducted using a single \emph{NVIDIA} A40 GPU. For the vision task, we implemented a self-designed single-head~\emph{Vision Transformer} (ViT)~\citep{vit} on \emph{CIFAR-10} and \emph{CIFAR-100} datasets~\citep{cifar} to enable rapid assessment of the feasibility of replacing \emph{LayerNorm} and \emph{Softmax} with LENS. We then extended the evaluation to \emph{Data-Efficient Image Transformer} (DeiT)~\citep{deit} with \emph{ImageNet-1K}~\citep{imagenet} classification. Moreover, we evaluated LENS using~\emph{Pythia}-14M/31M/70M~\citep{pythia} to test whether this architecture can be generalized to causal language models, rather than being limited to vision tasks. To simulate deploying the pretrained network models on the specifically designed inference hardware, such as a \emph{Neural Processing Unit} (NPU) or FPGA accelerators, we always choose or train a baseline in our experiments and then explore whether our LENS architecture can recover the accuracy with retraining after replacing \emph{LayerNorm} and \emph{Softmax}. If LENS can recover most of the baseline performance, it means that shifting most of the computation of \emph{LayerNorm} and \emph{Softmax} to the matrix-multiplication core and eliminating their corresponding hardware modules is possible in hardware design.

\subsection{CIFAR-10 and CIFAR-100 using Single-Head ViT}

We first conducted proof-of-concept experiments on \emph{CIFAR-10} and \emph{CIFAR-100} using a lightweight single-head \emph{ViT}. We input 65 tokens, each with a hidden dimension of 128, to the model. 64 tokens are generated from the $32\times32$-pixel RGB images with a $4\times4$ convolution kernel and a stride of 4. One additional \emph{Classification} (CLS) token is adopted, and learnable absolute positional embeddings are added to all tokens. The \emph{Transformer} backbone of the single-head \emph{ViT} consists of three \emph{Pre-Layer Normalization} (Pre-LN) blocks. Each block uses single-head scaled dot-product self-attention and a \emph{Feed Forward Network} (FFN) with dimensions 128, 512, and 128. The FFN applies the \emph{GELU} activation function. Both the attention and FFN sublayers use residual connections. The output projection and FFN layers use a dropout rate of 0.1, while the attention weights use no dropout. A dropout rate of 0.1 is also applied after adding the positional embeddings. After a final normalization layer, the CLS representation is passed to a linear classifier with 10 or 100 outputs.

\begin{table}[t]
    \centering
    \caption{Experimental Results on \emph{CIFAR-10} and \emph{CIFAR-100} using a Single-Head ViT with~\emph{LayerNorm},~\emph{Softmax},~\emph{LENSNorm}
    and~\emph{LENSMax}.}
    \vspace{5pt}
    \label{cifar_results}
    \resizebox{1\columnwidth}{!}
    {
        \begin{tabular}{cccccccc}
            \toprule
            Dataset                    & Training Path       & Epochs               & Peak LR                             & Norm.     & Attn.   & Top-1 Acc. ($\%$) & $\Delta$ Acc. (pp) \\ \midrule
            \multirow{8}{*}{CIFAR-10}  & \multirow{4}{*}{From Scratch}     & \multirow{4}{*}{400} & \multirow{4}{*}{$1 \times 10^{-3}$} & LayerNorm & Softmax & $80.33\pm0.28$    & $0.00$ (Baseline)    \\
                                       &                             &                      &                                     & LENSNorm  & Softmax & $80.18\pm0.80$    & $-0.15$              \\
                                       &                             &                      &                                     & LayerNorm & LENSMax & $78.82\pm0.43$    & $-1.51$              \\
                                       &                             &                      &                                     & LENSNorm  & LENSMax & $79.97\pm0.79$    & $-0.36$              \\ \cmidrule(lr){2-8}
                                       & Only-LENS                   & 50                   & $1 \times 10^{-3}$                  & LENSNorm  & LENSMax & $68.26\pm3.38$    & $-12.07$ \\ \cmidrule(lr){2-8}
                                       & \multirow{3}{*}{Scratch$\rightarrow$LENS} & \multirow{3}{*}{50}  & \multirow{3}{*}{$5 \times 10^{-4}$} & LENSNorm  & Softmax & $81.41\pm0.17$    & $+1.08$              \\
                                       &                             &                      &                                     & LayerNorm & LENSMax & $80.06\pm0.38$    & $-0.27$              \\
                                       &                             &                      &                                     & LENSNorm  & LENSMax & $81.00\pm0.19$    & $+0.67$              \\ \midrule[1.5pt]
            \multirow{8}{*}{CIFAR-100} & \multirow{4}{*}{From Scratch}     & \multirow{4}{*}{400} & \multirow{4}{*}{$1 \times 10^{-3}$} & LayerNorm & Softmax & $51.22\pm0.55$    & $0.00$ (Baseline)    \\
                                       &                             &                      &                                     & LENSNorm  & Softmax & $52.80\pm1.28$    & $+1.58$              \\
                                       &                             &                      &                                     & LayerNorm & LENSMax & $50.63\pm0.71$    & $-0.59$              \\
                                       &                             &                      &                                     & LENSNorm  & LENSMax & $53.03\pm1.44$    & $+1.81$              \\ \cmidrule(lr){2-8}
                                       & Only-LENS                   & 50                   & $1 \times 10^{-3}$                  & LENSNorm  & LENSMax & $38.50\pm1.70$    & $-12.72$ \\ \cmidrule(lr){2-8}
                                       & \multirow{3}{*}{Scratch$\rightarrow$LENS} & \multirow{3}{*}{50}  & \multirow{3}{*}{$5 \times 10^{-4}$} & LENSNorm  & Softmax & $52.14\pm0.83$    & $+0.92$              \\
                                       &                             &                      &                                     & LayerNorm & LENSMax & $51.01\pm0.72$    & $-0.21$              \\
                                       &                             &                      &                                     & LENSNorm  & LENSMax & $52.05\pm0.60$    & $+0.83$              \\ \bottomrule
        \end{tabular}
    }

    \vspace{5pt}
    \parbox{\linewidth}{\footnotesize \textbf{Note:} \textit{Norm.} and \textit{Attn.} denote the normalization and attention methods, respectively. \textit{From Scratch} denotes training from scratch. \textit{Scratch$\rightarrow$LENS} denotes retraining the indicated LENS configuration from the corresponding \textit{From Scratch} trained model. \textit{Only-LENS} denotes initialization from the corresponding \textit{From Scratch} trained model, with all original model parameters frozen and only the LENSNorm and LENSMax parameters trained. The \textit{From Scratch} models using both \emph{LayerNorm} and \emph{Softmax} serve as the baselines.}

    \vspace{-10pt}
    
\end{table}

As shown in~\autoref{cifar_results}, we trained the baseline single-head ViTs for \emph{CIFAR-10} and \emph{CIFAR-100} from scratch for 400 epochs using \emph{LayerNorm} and \emph{Softmax}. We then replaced \emph{LayerNorm}, \emph{Softmax}, or both with the corresponding LENS modules. We evaluated these configurations either by training from scratch or by initializing them from the corresponding trained baseline models and retraining all model parameters for 50 epochs. Furthermore, in the \textit{Only-LENS} setting, we froze all loaded baseline parameters and trained only \emph{LENSNorm} and \emph{LENSMax} MLPs for 50 epochs. All models were trained with a batch size of 256 using the \emph{AdamW}~\citep{adamw} optimizer with a weight decay of 0.05 and a cosine annealing learning rate scheduler. Results are reported as mean $\pm$ standard deviation over five random seeds, $\{42, 886, 1024, 1991, 2333\}$. We applied \emph{RandomCrop} with a padding of 4 and \emph{RandomHorizontalFlip} for data augmentation.

Therefore, based on the experimental results, we find that LENS can replace both \emph{LayerNorm} and \emph{Softmax} and can be trained from scratch. When both \emph{LayerNorm} and \emph{Softmax} are replaced, the \emph{From Scratch} models achieve accuracy close to the baseline on \emph{CIFAR-10} and exceed it on \emph{CIFAR-100} after 400 epochs. Initializing from the trained baselines enables faster recovery in only 50 epochs and exceeds the baselines by $0.67$ and $0.83$ percentage points, respectively. \emph{LENSNorm} generally performs better than \emph{LENSMax}. When we only replace \emph{LayerNorm} with \emph{LENSNorm}, the replaced model exceeds the baseline on both datasets after retraining. In contrast to \emph{LENSNorm}, only replacing \emph{Softmax} with \emph{LENSMax} causes small accuracy losses. Moreover, if we freeze the loaded model parameters and train only the LENS MLPs, accuracy loss is substantially larger than in other settings. This suggests that LENS does not simply approximate \emph{LayerNorm} and \emph{Softmax}, but jointly adapts with the original model parameters during retraining. In addition, we also explored the effects of replacing the \emph{DyT}+\emph{Softmax}, \emph{Derf}+\emph{Softmax}, and \emph{RMSNorm}+\emph{Softmax} with our \emph{LENSNorm} and \emph{LENSMax} in the experiments of CIFAR-10/100. Considering that the mechanism of our \emph{LENSMax} is abstracted from \emph{Softmax}, we have not adopted other alternatives to \emph{Softmax} in these experiments. As shown in \mbox{\autoref{additional_results}} of the Appendix.~\mbox{\ref{additional_result_section}}, experimental results show that the accuracy recovery behavior after replacing these normalization operations with LENS is consistent with that observed in our experiments with \emph{LayerNorm} and \emph{Softmax}.

\subsection{ImageNet using DeiT-Tiny and DeiT-Small}

\begin{table}[t]
    \centering
    \caption{Experimental Results on \emph{ImageNet} using DeiT-Tiny and DeiT-Small with~\emph{LayerNorm},~\emph{Softmax},~\emph{LENSNorm}
    and~\emph{LENSMax}.}
    \vspace{5pt}
    \label{imagenet_results}
    \resizebox{1\columnwidth}{!}
    {
        \begin{tabular}{cccccccc}
            \toprule
            \multirow{2}{*}{Model}      & \multirow{2}{*}{Training Path} & LR & Knowledge             & \multicolumn{2}{c}{Top-1}           & \multicolumn{2}{c}{Top-5}           \\
                                        &                           & Schedule              & Distillation          & Acc. ($\%$)    & $\Delta$ Acc. (pp) & Acc. ($\%$)    & $\Delta$ Acc. (pp) \\ \midrule
            DeiT-Tiny (Base.)        & Official    & -                            & -                     & $72.21$        & $0.00$             & $91.10$        & $0.00$             \\ \midrule
            \multirow{4}{*}{DeiT-Tiny+LENS}  & From Scratch          & Cos.             & No                    & $51.67$        & $-20.54$           & $75.29$        & $-15.81$           \\
                                        & Official$\rightarrow$LENS & Cos.             & No                    & $71.94$        & $-0.27$            & $90.53$        & $-0.57$            \\
                                        & Official$\rightarrow$LENS & Exp.                  & No                    & $72.44\pm0.03$ & $+0.23$            & $90.95\pm0.05$ & $-0.15$            \\
                                        & Official$\rightarrow$LENS & Exp.                  & Yes                   & $73.66\pm 0.26$ & $+1.45$            & $91.91\pm0.11$ & $+0.81$            \\ \midrule[1.5pt]
            DeiT-Small (Base.)       & Official                  & -                            & -                     & $79.87$        & $0.00$             & $95.06$        & $0.00$             \\ \midrule
            \multirow{2}{*}{DeiT-Small+LENS}  & Official$\rightarrow$LENS & Exp.                  & No                    & $74.51$        & $-5.36$             & $92.07$        & $-2.99$             \\
                                        & Official$\rightarrow$LENS & Exp.                  & Yes                   & $78.30$        & $-1.57$            & $94.12$        & $-0.94$             \\ \bottomrule
        \end{tabular}
    }

    \vspace{5pt}
    \parbox{\linewidth}{\footnotesize \textbf{Note:} \textit{Official} denotes the downloaded official pretrained DeiT model using \emph{LayerNorm} and \emph{Softmax}. \textit{From Scratch} and \textit{Official$\rightarrow$LENS} denote LENS models trained from scratch and initialized from the downloaded official pretrained model, respectively. All LENS models use both \emph{LENSNorm} and \emph{LENSMax}. \textit{Base.} denotes the baseline. Results using DeiT-Tiny with the exponential-decay learning-rate schedule are reported as mean $\pm$ standard deviation over three seeds; all other results are from single runs. \textit{Cos.} and \textit{Exp.} denote the cosine annealing and exponential decay learning-rate schedules.}

    \vspace{-10pt}
    
\end{table}

To evaluate the performance of LENS on larger ViT models and larger datasets, we further conducted experiments of LENS on DeiT-Tiny and DeiT-Small using ImageNet-1k. As shown in~\autoref{imagenet_results}, we downloaded the official checkpoints of DeiT-Tiny and DeiT-Small as baselines, and then replaced all \emph{LayerNorm} and \emph{Softmax} with LENS. We retrained the LENS-replaced models for 50 epochs using cosine annealing or exponential-decay learning-rate schedules with or without \emph{Knowledge Distillation} (KD). For DeiT-Tiny, we additionally explored training from scratch after replacing \emph{LayerNorm} and \emph{Softmax} with the LENS modules. The training results of DeiT-Tiny models using exponential-decay learning-rate schedules are reported as the mean $\pm$ standard deviation across three random seeds, $\{42, 1024, 2333\}$. All experiments were trained with a batch size of 256 and a peak learning rate of $7.5\times10^{-5}$ using the \emph{AdamW} optimizer with a weight decay of 0.05. The exponential decay rates $\gamma$ in DeiT-Tiny and DeiT-Small are $0.90$ and $0.95$, respectively. For the KD setting used in related experiments, we applied KD only to the output logits using the corresponding official checkpoint as the teacher, with a temperature $T=2$ and a weight $\alpha=0.5$. We followed the standard ImageNet data preprocessing and augmentation settings of the official DeiT implementation.

According to the experimental results in~\autoref{imagenet_results}, under the same 50-epoch training budget, training the LENS-replaced DeiT-Tiny from scratch leads to a large loss in accuracy, whereas initializing from the official pretrained model significantly improves accuracy recovery. Compared with cosine annealing, the exponential-decay learning-rate schedule performed better in our DeiT-Tiny experiments. With exponential decay, both DeiT-Tiny experiments with and without KD exceed the Top-1 accuracy of the baseline. The KD setting exceeds the baseline by $1.45$ and $0.81$ percentage points in Top-1 and Top-5 accuracy, respectively. For DeiT-Small, after applying KD and the exponential-decay learning-rate schedule, the LENS-replaced model also recovers most of the baseline accuracy, with losses of only $1.57$ and $0.94$ percentage points in Top-1 and Top-5 accuracy, respectively.

Moreover, to clarify whether the accuracy improvement in our DeiT-Tiny experiments comes from the LENS architecture or the additional 50 epochs of training, we apply the same setting with an exponential decay learning-rate schedule and the same peak and minimum learning rates to continuously train the official checkpoint of DeiT-Tiny for 50 epochs. The final accuracy is Top-1 $74.30\%\pm0.04\%$ and Top-5 $92.06\%\pm0.01\%$, which exceeds our experimental results of LENS. This experiment shows that LENS may not offer an accuracy advantage over regular \emph{LayerNorm} and \emph{Softmax}. However, improving accuracy is not the design goal of LENS: Shifting more arithmetic operations in \emph{Transformer} to matrix computation cores and eliminating the corresponding specific hardware of \emph{LayerNorm} and \emph{Softmax} while maintaining most of the network performance from the pretrained models. Considering DeiT baselines as the pretrained model to be deployed on hardware, current experimental results support that LENS can convert their \emph{LayerNorm} and \emph{Softmax} computations into MLPs and recover their accuracy with sufficient retraining.

\subsection{Language Modeling using Pythia-14M/31M/70M}

To evaluate whether LENS can be adopted in causal language modeling tasks rather than only vision models, we conducted proof-of-concept experiments using the \emph{Pythia}-14M/31M/70M models~\citep{pythia}. \autoref{harness_results} reports the validation and generalization results of the official \emph{Pythia}-14M checkpoint trained on \emph{The Pile}~\citep{pile}, its LENS-replaced version retrained on \emph{The Pile Subset}, and the \emph{Pythia}-14M/31M/70M models trained by us on \emph{FineWeb-Edu-3B}~\citep{fineweb}. All models were evaluated on \emph{WikiText}~\citep{wikitext} and \emph{LAMBADA}~\citep{lambada} using the \emph{lm-evaluation-harness} toolkit~\citep{eval-harness}. All models trained on \emph{FineWeb-Edu-3B} used a batch size of 16, the \emph{AdamW} optimizer with a weight decay of 0.01, and a cosine-annealing learning-rate schedule. The maximum sequence length was fixed at 512. For causal attention, future positions were masked before \emph{LENSMax}, and their output weights were set to zero, while $N$ was set to the number of valid elements in each row.

We first replaced all \emph{LayerNorm} and \emph{Softmax} in the official \emph{Pythia}-14M checkpoint with LENS and retrained the model for two epochs on a subset of \emph{The Pile}, containing 624000 training samples and 5000 validation samples. Immediately after replacement, the model performance dropped to near-random levels. We also observed the same situation in CIFAR-10/100 and DeiT experiments. As shown by the validation trajectories in~\mbox{\autoref{val_in_training}} of Appendix~\mbox{\ref{training_plot}}, its performance was gradually recovered during retraining. This indicates that replacing \emph{LayerNorm} and \emph{Softmax} with LENS has significantly destroyed the language ability of the original model. LENS replacement is not conventional fine-tuning or continuous training from the pretrained checkpoint, but joint retraining initialized from pretrained model parameters. Although the retrained model largely recovered its validation performance on \emph{The Pile Subset}, its performance collapsed on \emph{WikiText} and \emph{LAMBADA}. This suggests that retraining on the limited subset adapted the model to that subset but could not recover the broad language knowledge from a crushed model, whose architecture has been significantly changed while the replacement of LENS, because the language ability of the original model has been learned from the complete \emph{Pile} dataset. This failure motivated us to train the baseline and LENS models on the same \emph{FineWeb-Edu-3B} dataset. We trained the \emph{Pythia}-14M/31M/70M models from scratch using the \emph{FineWeb-Edu-3B} dataset as our baseline, simulating deployment on hardware. Then, we retrained the LENS-replaced model to recover the language performance with the complete \emph{FineWeb-Edu-3B} dataset.
 
\begin{table}[t]
    \centering
    \caption{Validation and Generalization Results of Pythia-14M/31M/70M using~\emph{LayerNorm},~\emph{Softmax},~\emph{LENSNorm}, and~\emph{LENSMax}.}
    \vspace{5pt}
    \label{harness_results}
    \resizebox{1\columnwidth}{!}
    {
        \begin{tabular}{cccclccclclcccl}
            \toprule
            \multicolumn{15}{c}{\textbf{Validation Results}}                                                                                                                                                                                                                                                                                                                                            \\ \midrule[2pt]
            Pythia & Training        & Training                  & \multicolumn{2}{c}{\multirow{2}{*}{Epochs}} & \multicolumn{2}{c}{LR}                    & \multicolumn{2}{c}{\multirow{2}{*}{PPL $\downarrow$}} & \multicolumn{2}{c}{\multirow{2}{*}{$\Delta$ PPL}} & \multicolumn{4}{c}{Next-Token}                                                      \\
            Model                &  Dataset                    & Path                      & \multicolumn{2}{c}{}                        & \multicolumn{2}{c}{Schedule}              & \multicolumn{2}{c}{}                                  & \multicolumn{2}{c}{}                              & \multicolumn{2}{c}{Acc. ($\%$) $\uparrow$} & \multicolumn{2}{c}{$\Delta$ Acc. (pp)} \\ \midrule
            14M (Base.)       & The Pile    & Official                  & \multicolumn{2}{c}{-}                       & \multicolumn{2}{c}{-}                     & \multicolumn{2}{c}{98.01}                             & \multicolumn{2}{c}{$0.00$}                        & \multicolumn{2}{c}{28.95}                  & \multicolumn{2}{c}{$0.00$}             \\
            14M+LENS        & The Pile Subset        & Official$\rightarrow$LENS & \multicolumn{2}{c}{2}                       & \multicolumn{2}{c}{Step.}                 & \multicolumn{2}{c}{92.54}                             & \multicolumn{2}{c}{$-5.47$}                       & \multicolumn{2}{c}{27.45}                  & \multicolumn{2}{c}{$-1.50$}            \\ \midrule
            14M (Base.)     & \multirow{3}{*}{FineWeb-Edu-3B} & From Scratch              & \multicolumn{2}{c}{\multirow{3}{*}{1}}      & \multicolumn{2}{c}{\multirow{3}{*}{Cos.}} & \multicolumn{2}{c}{56.67}                             & \multicolumn{2}{c}{$0.00$}                        & \multicolumn{2}{c}{31.56}                  & \multicolumn{2}{c}{$0.00$}             \\
            14M+LENS        &                                 & From Scratch              & \multicolumn{2}{c}{}                        & \multicolumn{2}{c}{}                      & \multicolumn{2}{c}{64.70}                             & \multicolumn{2}{c}{$+8.03$}                       & \multicolumn{2}{c}{30.13}                  & \multicolumn{2}{c}{$-1.43$}            \\
            14M+LENS        &                                 & Scratch$\rightarrow$LENS  & \multicolumn{2}{c}{}                        & \multicolumn{2}{c}{}                      & \multicolumn{2}{c}{$57.27\pm0.26$}                    & \multicolumn{2}{c}{$+0.60$}                       & \multicolumn{2}{c}{$31.35\pm0.07$}         & \multicolumn{2}{c}{$-0.21$}            \\ \midrule
            31M (Base.)     & \multirow{2}{*}{FineWeb-Edu-3B} & From Scratch              & \multicolumn{2}{c}{\multirow{2}{*}{1}}      & \multicolumn{2}{c}{\multirow{2}{*}{Cos.}} & \multicolumn{2}{c}{39.89}                             & \multicolumn{2}{c}{$0.00$}                        & \multicolumn{2}{c}{34.72}                  & \multicolumn{2}{c}{$0.00$}             \\
            31M+LENS        &                                 & Scratch$\rightarrow$LENS  & \multicolumn{2}{c}{}                        & \multicolumn{2}{c}{}                      & \multicolumn{2}{c}{40.77}                             & \multicolumn{2}{c}{$+0.88$}                       & \multicolumn{2}{c}{34.38}                  & \multicolumn{2}{c}{$-0.34$}            \\ \midrule
            70M (Base.)     & \multirow{2}{*}{FineWeb-Edu-3B} & From Scratch              & \multicolumn{2}{c}{\multirow{2}{*}{1}}      & \multicolumn{2}{c}{\multirow{2}{*}{Cos.}} & \multicolumn{2}{c}{30.45}                             & \multicolumn{2}{c}{$0.00$}                        & \multicolumn{2}{c}{37.36}                  & \multicolumn{2}{c}{$0.00$}             \\
            70M+LENS        &                                 & Scratch$\rightarrow$LENS  & \multicolumn{2}{c}{}                        & \multicolumn{2}{c}{}                      & \multicolumn{2}{c}{31.00}                             & \multicolumn{2}{c}{$+0.55$}                       & \multicolumn{2}{c}{37.03}                  & \multicolumn{2}{c}{$-0.33$}            \\  \midrule[2pt]
            \multicolumn{15}{c}{\textbf{Generalization Evaluation Results}}                                                                                                                                                                                                                                                                                                                           \\  \midrule[2pt]
            Pythia & Training        & Training                  & \multicolumn{12}{c}{WikiText}                                                                                                                                                                                                                                                             \\
            Model                 &  Dataset                       & Path                      & \multicolumn{3}{c}{PPL $\downarrow$}                               & \multicolumn{3}{c}{$\Delta$ PPL}                                           & \multicolumn{3}{c}{BPB $\downarrow$}                                     & \multicolumn{3}{c}{$\Delta$ BPB}                             \\ \midrule
            14M (Base.)     & The Pile    & Official                  & \multicolumn{3}{c}{$154.28$}                                       & \multicolumn{3}{c}{$0.00$}                                                 & \multicolumn{3}{c}{$1.359$}                                              & \multicolumn{3}{c}{$0.000$}                                  \\
            14M+LENS        &  The Pile Subset        & Official$\rightarrow$LENS & \multicolumn{3}{c}{$632.81$}                                       & \multicolumn{3}{c}{$+478.53$}                                              & \multicolumn{3}{c}{$1.740$}                                              & \multicolumn{3}{c}{$+0.381$}                                 \\ \midrule
            14M (Base.)     & \multirow{3}{*}{FineWeb-Edu-3B} & From Scratch              & \multicolumn{3}{c}{$158.71$}                                       & \multicolumn{3}{c}{$0.00$}                                                 & \multicolumn{3}{c}{$1.367$}                                              & \multicolumn{3}{c}{$0.000$}                                  \\
            14M+LENS        &                                 & From Scratch              & \multicolumn{3}{c}{$184.55$}                                       & \multicolumn{3}{c}{$+25.84$}                                               & \multicolumn{3}{c}{$1.408$}                                              & \multicolumn{3}{c}{$+0.041$}                                 \\
            14M+LENS        &                                 & Scratch$\rightarrow$LENS  & \multicolumn{3}{c}{$160.14\pm0.84$}                                & \multicolumn{3}{c}{$+1.43$}                                                & \multicolumn{3}{c}{$1.370\pm0.002$}                                      & \multicolumn{3}{c}{$+0.003$}                                 \\ \midrule
            31M (Base.)     & \multirow{2}{*}{FineWeb-Edu-3B} & From Scratch              & \multicolumn{3}{c}{$103.92$}                                       & \multicolumn{3}{c}{$0.00$}                                                 & \multicolumn{3}{c}{$1.253$}                                              & \multicolumn{3}{c}{$0.00$}                                   \\
            31M+LENS        &                                 & Scratch$\rightarrow$LENS  & \multicolumn{3}{c}{$107.29$}                                       & \multicolumn{3}{c}{$+3.37$}                                                & \multicolumn{3}{c}{$1.261$}                                              & \multicolumn{3}{c}{$+0.008$}                                 \\ \midrule
            70M (Base.)     & \multirow{2}{*}{FineWeb-Edu-3B} & From Scratch              & \multicolumn{3}{c}{$74.14$}                                        & \multicolumn{3}{c}{$0.00$}                                                 & \multicolumn{3}{c}{$1.162$}                                              & \multicolumn{3}{c}{$0.00$}                                   \\
            70M+LENS        &                                 & Scratch$\rightarrow$LENS  & \multicolumn{3}{c}{$76.33$}                                        & \multicolumn{3}{c}{$+2.19$}                                                & \multicolumn{3}{c}{$1.170$}                                              & \multicolumn{3}{c}{$+0.008$}                                 \\ \midrule[1.5pt]
            Pythia & Training        & Training                  & \multicolumn{12}{c}{LAMBADA}                                                                                                                                                                                                                                                              \\
            Model                 & Dataset                         & Path                      & \multicolumn{3}{c}{PPL $\downarrow$}                               & \multicolumn{3}{c}{$\Delta$ PPL}                                           & \multicolumn{3}{c}{Acc. ($\%$) $\uparrow$}                               & \multicolumn{3}{c}{$\Delta$ Acc. (pp)}                       \\ \midrule
            14M (Base.)     & The Pile    & Official                  & \multicolumn{3}{c}{$914.80\pm208.72$}                              & \multicolumn{3}{c}{$0.00$}                                                 & \multicolumn{3}{c}{$12.00\pm2.30$}                                       & \multicolumn{3}{c}{$0.00$}                                   \\
            14M+LENS        & The Pile Subset        & Official$\rightarrow$LENS & \multicolumn{3}{c}{$27028.53\pm6481.85$}                           & \multicolumn{3}{c}{$+26113.73$}                                            & \multicolumn{3}{c}{$0.00\pm0.00$}                                       & \multicolumn{3}{c}{$-12.00$}                                 \\ \midrule
            14M (Base.)     & \multirow{3}{*}{FineWeb-Edu-3B} & From Scratch              & \multicolumn{3}{c}{$3426.19\pm179.30$}                             & \multicolumn{3}{c}{$0.00$}                                                 & \multicolumn{3}{c}{$8.64\pm0.39$}                                        & \multicolumn{3}{c}{$0.00$}                                   \\
            14M+LENS        &                                 & From Scratch              & \multicolumn{3}{c}{$5484.52\pm272.83$}                             & \multicolumn{3}{c}{$+2058.33$}                                             & \multicolumn{3}{c}{$4.70\pm0.29$}                                        & \multicolumn{3}{c}{$-3.94$}                                  \\
            14M+LENS        &                                 & Scratch$\rightarrow$LENS  & \multicolumn{3}{c}{$3587.40\pm349.60$}                             & \multicolumn{3}{c}{$+161.21$}                                              & \multicolumn{3}{c}{$6.99\pm0.58$}                                        & \multicolumn{3}{c}{$-1.65$}                                  \\ \midrule
            31M (Base.)     & \multirow{2}{*}{FineWeb-Edu-3B} & From Scratch              & \multicolumn{3}{c}{$830.77\pm42.35$}                               & \multicolumn{3}{c}{$0.00$}                                                 & \multicolumn{3}{c}{$14.94\pm0.50$}                                       & \multicolumn{3}{c}{$0.00$}                                   \\
            31M+LENS        &                                 & Scratch$\rightarrow$LENS  & \multicolumn{3}{c}{$875.69\pm43.99$}                               & \multicolumn{3}{c}{$+44.92$}                                               & \multicolumn{3}{c}{$12.83\pm0.47$}                                       & \multicolumn{3}{c}{$-2.11$}                                  \\ \midrule
            70M (Base.)     & \multirow{2}{*}{FineWeb-Edu-3B} & From Scratch              & \multicolumn{3}{c}{$346.34\pm17.14$}                               & \multicolumn{3}{c}{$0.00$}                                                 & \multicolumn{3}{c}{$18.32\pm0.54$}                                       & \multicolumn{3}{c}{$0.00$}                                   \\
            70M+LENS        &                                 & Scratch$\rightarrow$LENS  & \multicolumn{3}{c}{$304.73\pm14.80$}                               & \multicolumn{3}{c}{$-41.61$}                                               & \multicolumn{3}{c}{$18.09\pm0.54$}                                       & \multicolumn{3}{c}{$-0.23$}                                  \\ \bottomrule
        \end{tabular}
    }

    \vspace{5pt}
    \parbox{\linewidth}{\footnotesize \textbf{Note:} \textit{Official} is the released pretrained Pythia-14M checkpoint. \textit{Official$\rightarrow$LENS} and \textit{Scratch$\rightarrow$LENS} initialize LENS from this checkpoint and the corresponding baseline we trained. \textit{From Scratch} uses a random initialization. All LENS models contain both \emph{LENSNorm} and \emph{LENSMax}. \textit{Base.} uses \emph{LayerNorm} and \emph{Softmax}. The \emph{Pythia}-14M \textit{Scratch$\rightarrow$LENS} results on \emph{FineWeb-Edu-3B} are mean $\pm$ standard deviation over three training seeds; all others are single runs. For the single-run \emph{LAMBADA} evaluations, $\pm$ indicates the standard error reported by the evaluation harness. \textit{Cos.}/\textit{Step.} indicate cosine annealing/two-stage step decay. \textit{PPL}/\textit{BPB} indicate perplexity/bits per byte.}

    \vspace{-20pt}
    
\end{table}

For the \emph{FineWeb-Edu-3B} experiments, each training or retraining stage used one epoch. The learning rate was linearly warmed up over the first $1\%$ of training steps to $1\times10^{-3}$ and then cosine-annealed to $1\times10^{-4}$. Under the same one-epoch training budget, the Pythia-14M LENS model trained from random initialization performs worse than the standard \emph{LayerNorm}/\emph{Softmax} baseline on validation, \emph{WikiText}, and \emph{LAMBADA}, indicating that LENS is more difficult to train from scratch. However, when initialized from our trained baseline and retrained on the same dataset, the LENS models recover most of the baseline performance across Pythia-14M/31M/70M. Their validation next-token accuracy differs from the baselines by only $-0.21$, $-0.34$, and $-0.33$ percentage points, respectively. Their generalization performance on \emph{WikiText} and \emph{LAMBADA} also recovers substantially, with the 70M LENS model even achieving lower LAMBADA PPL than its baseline. These results suggest that LENS is applicable to causal language models, but its optimization is more difficult and benefits substantially from initialization with trained model parameters.

\section{Discussion}
\label{discussion}
We discuss the features and limitations of LENS through the following questions:

\textbf{Q1. What are the strengths of LENS on hardware design?}

Modern AI processors, such as NPUs, typically consist of matrix computation cores (such as the systolic array), specific arithmetic units (such as GELU, LayerNorm, Softmax, etc.), a state machine to control the computational progress, data input/output buffers, and so on. Considering that the manufacturing cost and power consumption are always directly proportional to the chip area of ASICs or the hardware resource consumption of FPGAs, if LENS can eliminate the corresponding hardware of \emph{LayerNorm} and \emph{Softmax} by shifting them to matrix computation cores, the associated costs can be reduced. LENS consists of fully connected linear layers, \emph{ReLU}, \emph{Sigmoid}, and power-of-two scaling operations. Linear layers can reuse the matrix computation core without additional hardware cost. \emph{ReLU} and power-of-two scaling operations are simple comparators and bitwise shifters (for fixed-point data) or integer subtraction (for floating-point data). \emph{Sigmoid} approximation based on thresholds and bitwise shifting is also a very mature and low-cost hardware design \mbox{\citep{pla1997}}. 

\textbf{Q2. Does LENS simply approximate \emph{LayerNorm} and \emph{Softmax}?}

\begin{figure}[t]
    \centering
    {
        \begin{minipage}{0.495\columnwidth}
            \begin{minipage}{0.495\columnwidth}
                \centerline{\includegraphics[width=1\columnwidth]{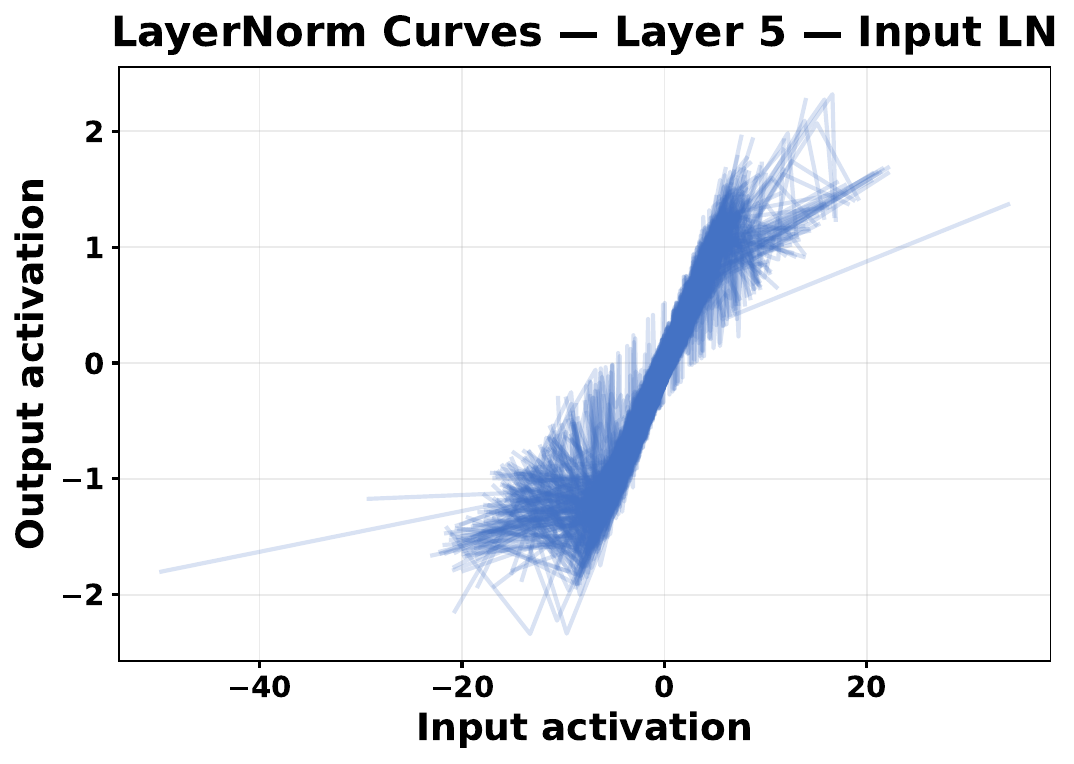}}
            \end{minipage}
            \begin{minipage}{0.495\columnwidth}
                \centerline{\includegraphics[width=1\columnwidth]{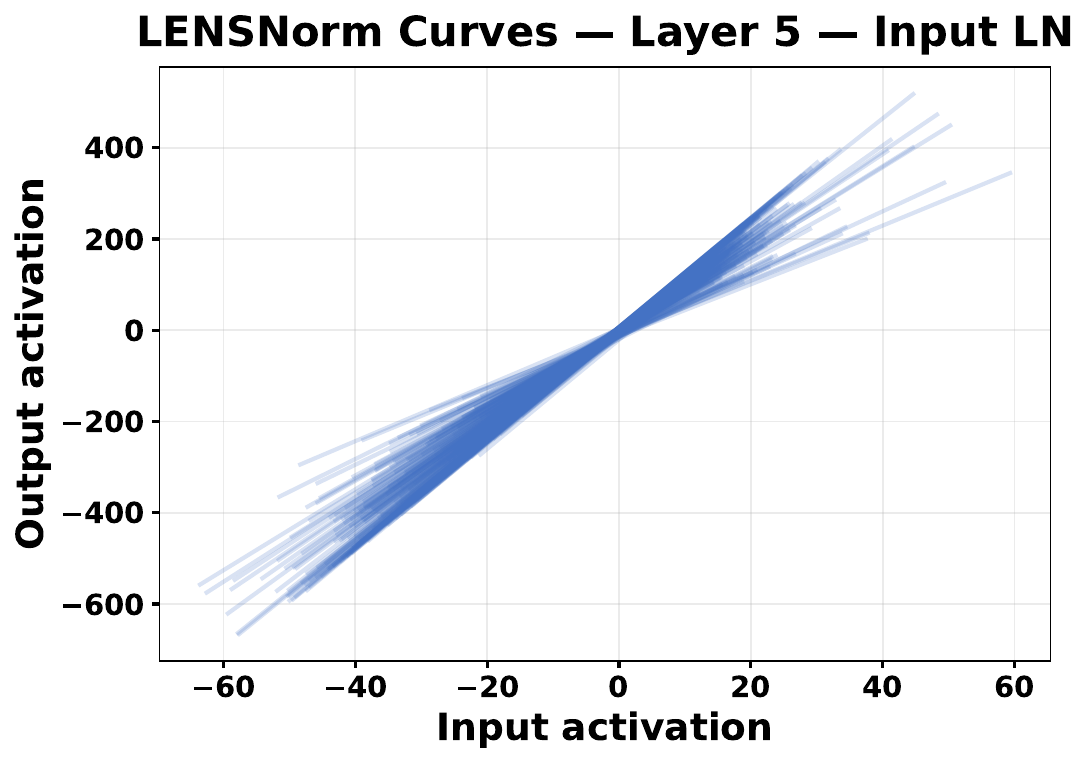}}
            \end{minipage}
            \centerline{\scriptsize (a) LayerNorm (left) and LENSNorm (right) in Pythia-14M}
        \end{minipage}
        \begin{minipage}{0.495\columnwidth}
            \begin{minipage}{0.495\columnwidth}
                \centerline{\includegraphics[width=1\columnwidth]{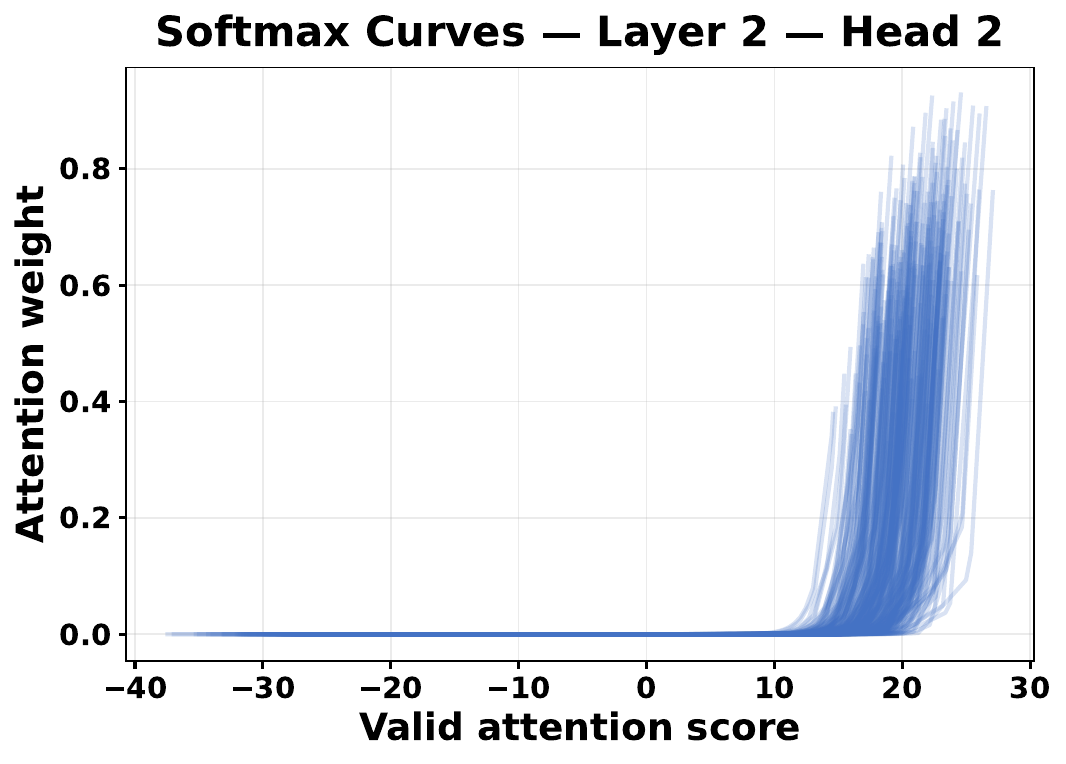}}
            \end{minipage}
            \begin{minipage}{0.495\columnwidth}
                \centerline{\includegraphics[width=1\columnwidth]{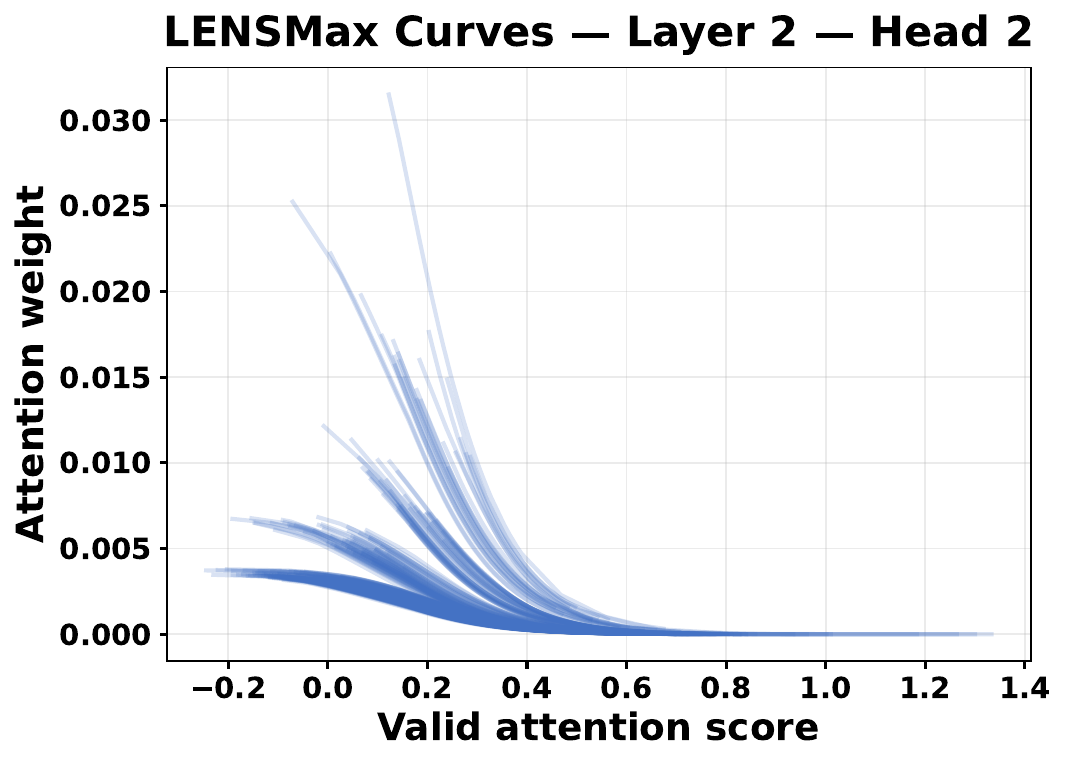}}
            \end{minipage}
            \centerline{\scriptsize (b) Softmax (left) and LENSMax (right) in Pythia-14M}
        \end{minipage}
    }
    \vspace{-10pt}
    \caption{Representative functional curves of LayerNorm and Softmax compared with LENS.}
    \label{softmax_vs_lensmax}
    \vspace{-15pt}
\end{figure}

No. As shown in~\autoref{softmax_vs_lensmax}, we plot representative functional curves of \emph{LayerNorm} and \emph{Softmax} and compare them with those produced by the corresponding LENS modules after retraining. For \emph{LayerNorm} and \emph{LENSNorm}, we evaluated 256 tokens to draw the functional curves. For \emph{Softmax} and \emph{LENSMax}, we collected 256 rows of attention scores. The results show that \emph{LENSNorm} and \emph{LENSMax} learn substantially different functional mappings from the original operators. In Appendix~\ref{relative_l2}, we further compare the relative $\ell_2$ weight changes between the baseline and retrained LENS models. The results show substantial changes in some model parameters during retraining. Together with the large accuracy loss in the \textit{Only-LENS} setting in the single-head ViT experiments, these observations suggest that LENS does not merely approximate the original operators but instead jointly adapts with the model parameters.

\textbf{Q3. What are the limitations of LENS?}

LENS is not a zero-shot replacement of \emph{LayerNorm} and \emph{Softmax}. Retraining is necessary and important for LENS. Some experimental results are not lossless compared with baselines. Moreover, the current \emph{LENSMax} requires a predefined maximum input length. Supporting length-agnostic inputs, evaluating larger models, and conducting direct hardware evaluation remain our future work. The current LENS research is a proof of concept showing that tiny learnable MLPs can serve as alternatives to the hand-designed \emph{LayerNorm} and \emph{Softmax}.

\section{Conclusion}
\label{conclusion}
In this work, we presented LENS, a GEMM-centric architecture that replaces all \emph{LayerNorm} and attention \emph{Softmax} operations in \emph{Transformers} with tiny learnable MLPs. Experiments on CIFAR-10/100, ImageNet-1K with DeiT, and causal language modeling with Pythia show that LENS can recover most baseline performance and even exceed it in several vision settings. Our analyses further show that LENS does not simply approximate the original operators but jointly adapts with the model parameters during retraining. These results provide a proof of concept that hand-designed \emph{LayerNorm} and \emph{Softmax} functions are not the only choices for \emph{Transformers} and motivate future research on scalable training and hardware implementation.

\subsection*{AI use statement}
\label{ai_use}



The research question, core mechanism, and principal methodological design of LENS were conceived and developed by the authors independently of generative AI. We used the web-based \emph{ChatGPT} interface as an interactive research assistant. No autonomous AI agent was used to design, execute, or manage the research workflow. The use of \emph{ChatGPT} was author-driven: the authors first formulated the intended research ideas, experimental designs, and implementation approaches, and then selectively considered suggestions provided by \emph{ChatGPT}. \emph{ChatGPT} was used to assist with boilerplate code for loading pretrained backbone models and preparing datasets, to discuss implementation and debugging issues, to review intermediate experimental diagnostics, and to improve the grammar, clarity, and readability of parts of the manuscript. AI-generated code suggestions were generally used as references rather than automatically incorporated into the implementation. The authors independently wrote or modified the final code based on their own design decisions after considering these suggestions. Any AI-assisted code used in the experiments was manually reviewed, tested, and verified by the authors before use. All research hypotheses, methodological contributions, experimental decisions, interpretation of the final results, and scientific claims were determined and verified by the authors. The authors take full responsibility for the final code, experiments, and manuscript.

\subsection*{Reproducibility statement}
\label{reproducibility}
The complete source code for all \emph{CIFAR-10} and \emph{CIFAR-100} experiments, including the full implementation of LENS, is provided at \href{https://github.com/liuyh-Horizon/LENS}{Github}. The code for the DeiT and Pythia experiments, together with an installable LENS package, will be publicly released upon acceptance.




\bibliography{iclr2027_conference}

@inproceedings{transformer,
    author = {Vaswani, Ashish and Shazeer, Noam and Parmar, Niki and Uszkoreit, Jakob and Jones, Llion and Gomez, Aidan N. and Kaiser, {\L}ukasz and Polosukhin, Illia},
    title = {Attention is all you need},
    year = {2017},
    isbn = {9781510860964},
    publisher = {Curran Associates Inc.},
    address = {Red Hook, NY, USA},
    booktitle = {Proceedings of the 31st International Conference on Neural Information Processing Systems},
    pages = {6000–6010},
    numpages = {11},
    location = {Long Beach, California, USA},
    series = {NIPS'17}
}

@misc{layernorm,
    title={Layer Normalization}, 
    author={Jimmy Lei Ba and Jamie Ryan Kiros and Geoffrey E. Hinton},
    year={2016},
    eprint={1607.06450},
    archivePrefix={arXiv},
    primaryClass={stat.ML},
    url={https://arxiv.org/abs/1607.06450}, 
}

@InProceedings{deit,
  title = 	 {Training data-efficient image transformers \& distillation through attention},
  author =       {Touvron, Hugo and Cord, Matthieu and Douze, Matthijs and Massa, Francisco and Sablayrolles, Alexandre and Jegou, Herve},
  booktitle = 	 {Proceedings of the 38th International Conference on Machine Learning},
  pages = 	 {10347--10357},
  year = 	 {2021},
  editor = 	 {Meila, Marina and Zhang, Tong},
  volume = 	 {139},
  series = 	 {Proceedings of Machine Learning Research},
  month = 	 {18--24 Jul},
  publisher =    {PMLR},
  url = 	 {https://proceedings.mlr.press/v139/touvron21a.html}
}

@inproceedings{
    vit,
    title={An Image is Worth 16x16 Words: Transformers for Image Recognition at Scale},
    author={Alexey Dosovitskiy and Lucas Beyer and Alexander Kolesnikov and Dirk Weissenborn and Xiaohua Zhai and Thomas Unterthiner and Mostafa Dehghani and Matthias Minderer and Georg Heigold and Sylvain Gelly and Jakob Uszkoreit and Neil Houlsby},
    booktitle={International Conference on Learning Representations},
    year={2021},
    url={https://openreview.net/forum?id=YicbFdNTTy}
}

@inproceedings{pythia,
author = {Biderman, Stella and Schoelkopf, Hailey and Anthony, Quentin and Bradley, Herbie and O'Brien, Kyle and Hallahan, Eric and Khan, Mohammad Aflah and Purohit, Shivanshu and Prashanth, USVSN Sai and Raff, Edward and Skowron, Aviya and Sutawika, Lintang and Van Der Wal, Oskar},
title = {Pythia: a suite for analyzing large language models across training and scaling},
year = {2023},
publisher = {JMLR.org},
booktitle = {Proceedings of the 40th International Conference on Machine Learning},
articleno = {102},
numpages = {34},
location = {Honolulu, Hawaii, USA},
series = {ICML'23}
}

@inbook{rmsnorm,
author = {Zhang, Biao and Sennrich, Rico},
title = {Root mean square layer normalization},
year = {2019},
publisher = {Curran Associates Inc.},
address = {Red Hook, NY, USA},
booktitle = {Proceedings of the 33rd International Conference on Neural Information Processing Systems},
articleno = {1110},
numpages = {12}
}

@InProceedings{dyt,
    author    = {Zhu, Jiachen and Chen, Xinlei and He, Kaiming and LeCun, Yann and Liu, Zhuang},
    title     = {Transformers without Normalization},
    booktitle = {Proceedings of the IEEE/CVF Conference on Computer Vision and Pattern Recognition (CVPR)},
    month     = {June},
    year      = {2025},
    pages     = {14901-14911}
}

@article{llm_fpga,
author = {Chen, Hongzheng and Zhang, Jiahao and Du, Yixiao and Xiang, Shaojie and Yue, Zichao and Zhang, Niansong and Cai, Yaohui and Zhang, Zhiru},
title = {Understanding the Potential of FPGA-based Spatial Acceleration for Large Language Model Inference},
year = {2024},
issue_date = {March 2025},
publisher = {Association for Computing Machinery},
address = {New York, NY, USA},
volume = {18},
number = {1},
issn = {1936-7406},
url = {https://doi.org/10.1145/3656177},
doi = {10.1145/3656177},
journal = {ACM Trans. Reconfigurable Technol. Syst.},
month = dec,
articleno = {5},
numpages = {29}
}

@inproceedings{bert,
    title = "{BERT}: Pre-training of Deep Bidirectional Transformers for Language Understanding",
    author = "Devlin, Jacob  and
      Chang, Ming-Wei  and
      Lee, Kenton  and
      Toutanova, Kristina",
    editor = "Burstein, Jill  and
      Doran, Christy  and
      Solorio, Thamar",
    booktitle = "Proceedings of the 2019 Conference of the North {A}merican Chapter of the Association for Computational Linguistics: Human Language Technologies, Volume 1 (Long and Short Papers)",
    month = jun,
    year = "2019",
    address = "Minneapolis, Minnesota",
    publisher = "Association for Computational Linguistics",
    url = "https://aclanthology.org/N19-1423/",
    doi = "10.18653/v1/N19-1423",
    pages = "4171--4186"
}

@article{gpt2,
  title={Language models are unsupervised multitask learners},
  author={Radford, Alec and Wu, Jeffrey and Child, Rewon and Luan, David and Amodei, Dario and Sutskever, Ilya and others},
  journal={OpenAI blog},
  volume={1},
  number={8},
  pages={9},
  year={2019}
}

@InProceedings{derf,
    author    = {Chen, Mingzhi and Lu, Taiming and Zhu, Jiachen and Sun, Mingjie and Liu, Zhuang},
    title     = {Stronger Normalization-Free Transformers},
    booktitle = {Proceedings of the IEEE/CVF Conference on Computer Vision and Pattern Recognition (CVPR)},
    month     = {June},
    year      = {2026},
    pages     = {27418-27428}
}

@inproceedings{
    seednorm,
    title={{S}ee{DN}orm: Self-Rescaled Dynamic Normalization},
    author={Wenrui Cai and Defa Zhu and Siyuan Qiao and Qingjie Liu and Qiyang Min},
    booktitle={The Fourteenth International Conference on Learning Representations},
    year={2026},
    url={https://openreview.net/forum?id=dFL7XExDvv}
}

@INPROCEEDINGS{
    sima,
    author={Koohpayegani, Soroush Abbasi and Pirsiavash, Hamed},
    booktitle={2024 IEEE/CVF Winter Conference on Applications of Computer Vision (WACV)}, 
    title={Sim{A}: Simple Softmax-free Attention for Vision Transformers}, 
    year={2024},
    volume={},
    number={},
    pages={2595-2605},
    doi={10.1109/WACV57701.2024.00259}
}

@inproceedings{ivit,
  title={{I-ViT}: Integer-only quantization for efficient vision transformer inference},
  author={Li, Zhikai and Gu, Qingyi},
  booktitle={Proceedings of the IEEE/CVF International Conference on Computer Vision},
  pages={17065--17075},
  year={2023}
}

@techreport{nvidia_blackwell,
  author      = {{NVIDIA Corporation}},
  title       = {{NVIDIA RTX Blackwell GPU Architecture: Built for Neural Rendering}},
  institution = {NVIDIA Corporation},
  type        = {White Paper},
  year        = {2025},
  note        = {Version 1.1},
  url         = {https://images.nvidia.com/aem-dam/Solutions/geforce/blackwell/nvidia-rtx-blackwell-gpu-architecture.pdf}
}

@misc{relu_attn,
  title={Replacing softmax with ReLU in Vision Transformers}, 
  author={Mitchell Wortsman and Jaehoon Lee and Justin Gilmer and Simon Kornblith},
  year={2023},
  eprint={2309.08586},
  archivePrefix={arXiv},
  primaryClass={cs.CV},
  url={https://arxiv.org/abs/2309.08586}, 
}

@inproceedings{soft,
    title={{SOFT}: Softmax-free Transformer with Linear Complexity}, 
    author={Lu, Jiachen and Yao, Jinghan and Zhang, Junge and Zhu, Xiatian and Xu, Hang and Gao, Weiguo and Xu, Chunjing and Xiang, Tao and Zhang, Li},
    booktitle={NeurIPS},
    year={2021}
}

@book{abramowitz1964handbook,
  title     = {Handbook of Mathematical Functions with Formulas, Graphs, and Mathematical Tables},
  editor    = {Milton Abramowitz and Irene A. Stegun},
  series    = {Applied Mathematics Series},
  volume    = {55},
  publisher = {National Bureau of Standards},
  address   = {Washington, DC},
  year      = {1964}
}

@misc{gelu,
      title={Gaussian Error Linear Units (GELUs)}, 
      author={Dan Hendrycks and Kevin Gimpel},
      year={2016},
      eprint={1606.08415},
      archivePrefix={arXiv},
      primaryClass={cs.LG},
      url={https://arxiv.org/abs/1606.08415}, 
}

@techreport{cifar,
  title       = {Learning multiple layers of features from tiny images},
  author      = {Krizhevsky, Alex},
  year        = {2009},
  institution = {University of Toronto}
}

@INPROCEEDINGS{imagenet,
  author={Deng, Jia and Dong, Wei and Socher, Richard and Li, Li-Jia and Kai Li and Li Fei-Fei},
  booktitle={2009 IEEE Conference on Computer Vision and Pattern Recognition}, 
  title={Image{N}et: A large-scale hierarchical image database}, 
  year={2009},
  volume={},
  number={},
  pages={248-255},
  doi={10.1109/CVPR.2009.5206848}}

@misc{pile,
      title={The {P}ile: An 800GB Dataset of Diverse Text for Language Modeling}, 
      author={Leo Gao and Stella Biderman and Sid Black and Laurence Golding and Travis Hoppe and Charles Foster and Jason Phang and Horace He and Anish Thite and Noa Nabeshima and Shawn Presser and Connor Leahy},
      year={2020},
      eprint={2101.00027},
      archivePrefix={arXiv},
      primaryClass={cs.CL},
      url={https://arxiv.org/abs/2101.00027}, 
}

@InProceedings{dn,
  title = 	 {Differentiable Dynamic Normalization for Learning Deep Representation},
  author =       {Luo, Ping and Zhanglin, Peng and Wenqi, Shao and Ruimao, Zhang and Jiamin, Ren and Lingyun, Wu},
  booktitle = 	 {Proceedings of the 36th International Conference on Machine Learning},
  pages = 	 {4203--4211},
  year = 	 {2019},
  editor = 	 {Chaudhuri, Kamalika and Salakhutdinov, Ruslan},
  volume = 	 {97},
  series = 	 {Proceedings of Machine Learning Research},
  month = 	 {09--15 Jun},
  publisher =    {PMLR},
  url = 	 {https://proceedings.mlr.press/v97/luo19a.html}
}

@INPROCEEDINGS{SAL-ViT,
  author={Zhang, Yuke and Chen, Dake and Kundu, Souvik and Li, Chenghao and Beerel, Peter A.},
  booktitle={2023 IEEE/CVF International Conference on Computer Vision (ICCV)}, 
  title={SAL-ViT: Towards Latency Efficient Private Inference on ViT using Selective Attention Search with a Learnable Softmax Approximation}, 
  year={2023},
  volume={},
  number={},
  pages={5093-5102},
  doi={10.1109/ICCV51070.2023.00472}}

@article{pla1997,
author = {H. Amin  and K.M. Curtis  and B.R. Hayes-Gill },
title = {Piecewise linear approximation
applied to nonlinear function of a neural network},
journal = {IEE Proceedings - Circuits, Devices and Systems},
volume = {144},
issue = {6},
pages = {313-317},
year = {1997},
doi = {10.1049/ip-cds:19971587},

URL = {https://digital-library.theiet.org/doi/abs/10.1049/ip-cds%3A19971587},
eprint = {https://digital-library.theiet.org/doi/pdf/10.1049/ip-cds%3A19971587}
}

@inproceedings{fineweb,
    author = {Penedo, Guilherme and Kydl{\'i}{\v c}ek, Hynek and Allal, Loubna Ben and Lozhkov, Anton and Mitchell, Margaret and Raffel, Colin and Von Werra, Leandro and Wolf, Thomas},
    title = {The FineWeb datasets: decanting the web for the finest text data at scale},
    year = {2024},
    isbn = {9798331314385},
    publisher = {Curran Associates Inc.},
    address = {Red Hook, NY, USA},
    booktitle = {Proceedings of the 38th International Conference on Neural Information Processing Systems},
    articleno = {970},
    numpages = {39},
    location = {Vancouver, BC, Canada},
    series = {NIPS '24}
}

@inproceedings{wikitext,
    title={Pointer Sentinel Mixture Models},
    author={Stephen Merity and Caiming Xiong and James Bradbury and Richard Socher},
    booktitle={International Conference on Learning Representations},
    year={2017},
    url={https://openreview.net/forum?id=Byj72udxe}
}

@inproceedings{lambada,
    title = "The {LAMBADA} dataset: Word prediction requiring a broad discourse context",
    author = "Paperno, Denis  and
      Kruszewski, Germ{\'a}n  and
      Lazaridou, Angeliki  and
      Pham, Ngoc Quan  and
      Bernardi, Raffaella  and
      Pezzelle, Sandro  and
      Baroni, Marco  and
      Boleda, Gemma  and
      Fern{\'a}ndez, Raquel",
    editor = "Erk, Katrin  and
      Smith, Noah A.",
    booktitle = "Proceedings of the 54th Annual Meeting of the Association for Computational Linguistics (Volume 1: Long Papers)",
    month = aug,
    year = "2016",
    address = "Berlin, Germany",
    publisher = "Association for Computational Linguistics",
    url = "https://aclanthology.org/P16-1144/",
    doi = "10.18653/v1/P16-1144",
    pages = "1525--1534"
}

@misc{eval-harness,
  author       = {Gao, Leo and Tow, Jonathan and Abbasi, Baber and Biderman, Stella and Black, Sid and DiPofi, Anthony and Foster, Charles and Golding, Laurence and Hsu, Jeffrey and Le Noac'h, Alain and Li, Haonan and McDonell, Kyle and Muennighoff, Niklas and Ociepa, Chris and Phang, Jason and Reynolds, Laria and Schoelkopf, Hailey and Skowron, Aviya and Sutawika, Lintang and Tang, Eric and Thite, Anish and Wang, Ben and Wang, Kevin and Zou, Andy},
  title        = {The Language Model Evaluation Harness},
  month        = 07,
  year         = 2024,
  publisher    = {Zenodo},
  version      = {v0.4.3},
  doi          = {10.5281/zenodo.12608602},
  url          = {https://zenodo.org/records/12608602}
}

@inproceedings{adamw,
    title={Decoupled Weight Decay Regularization},
    author={Ilya Loshchilov and Frank Hutter},
    booktitle={International Conference on Learning Representations},
    year={2019},
    url={https://openreview.net/forum?id=Bkg6RiCqY7},
}

@techreport{nvidia_fermi,
  author      = {{NVIDIA Corporation}},
  title       = {{NVIDIA's Next Generation CUDA Compute Architecture: Fermi}},
  institution = {NVIDIA Corporation},
  year        = {2009},
  type        = {White Paper},
  note        = {Version 1.1},
  url         = {https://www.nvidia.com/content/pdf/fermi_white_papers/nvidia_fermi_compute_architecture_whitepaper.pdf}
}
\bibliographystyle{iclr2027_conference}

\appendix
\section{Appendix}
\label{appendix}
\subsection{Unified Formulation of LayerNorm and Its Alternatives for LENSNorm}
\label{math_lens}

\subsubsection{LayerNorm}
\begin{equation}
\begin{aligned}
    \operatorname{LayerNorm}(\mathbf{x})
    &= \frac{\mathbf{x}-\operatorname{Mean}(\mathbf{x})}{\sqrt{\operatorname{Var}(\mathbf{x})+\epsilon}} \odot \boldsymbol{\gamma}+\boldsymbol{\beta} \\
    &= \frac{\boldsymbol{\gamma}}{\sqrt{\operatorname{Var}(\mathbf{x})+\epsilon}} \odot \mathbf{x} + \boldsymbol{\beta} - \frac{\operatorname{Mean}(\mathbf{x})\boldsymbol{\gamma}}{\sqrt{\operatorname{Var}(\mathbf{x})+\epsilon}} \\
    &= \mathbf{a}(\mathbf{x})\odot\mathbf{x} + \mathbf{b}(\mathbf{x}), \\[4pt]
    \mathbf{a}(\mathbf{x}) & = \frac{\boldsymbol{\gamma}}{\sqrt{\operatorname{Var}(\mathbf{x})+\epsilon}}, \\
    \qquad
    \mathbf{b}(\mathbf{x}) & = \frac{\sqrt{\operatorname{Var}(\mathbf{x})+\epsilon}\,\boldsymbol{\beta} - \operatorname{Mean}(\mathbf{x})\,\boldsymbol{\gamma}
    }{ \sqrt{\operatorname{Var}(\mathbf{x})+\epsilon} }.
\end{aligned}
\label{layernorm}
\end{equation}

\subsubsection{RMSNorm}

\begin{equation}
\begin{aligned}
    \operatorname{RMSNorm}(\mathbf{x})
    &= \frac{\mathbf{x}}
    {\sqrt{\operatorname{Mean}(\mathbf{x}\odot\mathbf{x})+\epsilon}}
    \odot \boldsymbol{\gamma} \\
    &= \frac{\boldsymbol{\gamma}}
    {\sqrt{\operatorname{Mean}(\mathbf{x}\odot\mathbf{x})+\epsilon}}
    \odot \mathbf{x} \\
    &= \mathbf{a}(\mathbf{x})\odot\mathbf{x}
    + \mathbf{b}(\mathbf{x}), \\[4pt]
    \mathbf{a}(\mathbf{x})
    &= \frac{\boldsymbol{\gamma}}
    {\sqrt{\operatorname{Mean}(\mathbf{x}\odot\mathbf{x})+\epsilon}}, \\
    \mathbf{b}(\mathbf{x})
    &= \mathbf{0}.
\end{aligned}
\label{rmsnorm}
\end{equation}

\subsubsection{DyT}

\begin{equation}
\begin{aligned}
    \operatorname{DyT}(\mathbf{x})
    &= \operatorname{tanh}(\alpha\mathbf{x})
    \odot \boldsymbol{\gamma}+\boldsymbol{\beta} \\
    &\approx
    \left(
    \alpha\mathbf{x}
    -\frac{\alpha^3}{3}\mathbf{x}^{\odot 3}
    \right)
    \odot \boldsymbol{\gamma}
    +\boldsymbol{\beta} \\
    &=
    \left(
    \alpha\boldsymbol{\gamma}
    -\frac{\alpha^3}{3}
    \boldsymbol{\gamma}\odot\mathbf{x}^{\odot 2}
    \right)
    \odot\mathbf{x}
    +\boldsymbol{\beta} \\
    &= \mathbf{a}(\mathbf{x})\odot\mathbf{x}
    + \mathbf{b}(\mathbf{x}), \\[4pt]
    \mathbf{a}(\mathbf{x})
    &= \alpha\boldsymbol{\gamma}
    -\frac{\alpha^3}{3}
    \boldsymbol{\gamma}\odot\mathbf{x}^{\odot 2}, \\
    \mathbf{b}(\mathbf{x})
    &= \boldsymbol{\beta}.
\end{aligned}
\label{dyt}
\end{equation}

\subsubsection{Derf}

The error function is approximated as~\citep{abramowitz1964handbook}
\begin{equation}
    \operatorname{erf}(\mathbf{x})
    \approx
    \operatorname{sign}(\mathbf{x})
    \left(
    1-
    \frac{1}{
    \left(
    1+a_1|\mathbf{x}|
    +a_2|\mathbf{x}|^2
    +a_3|\mathbf{x}|^3
    +a_4|\mathbf{x}|^4
    \right)^4}
    \right).
\end{equation}
Accordingly, $\mathcal{R}(z)$ in~\autoref{formular_compare} denotes
\begin{equation}
    \mathcal{R}(z)
    =
    \frac{
    1-
    \left(
    1+a_1|z|
    +a_2|z|^2
    +a_3|z|^3
    +a_4|z|^4
    \right)^{-4}
    }{|z|}.
\end{equation}

\begin{equation}
\begin{aligned}
    \operatorname{Derf}(\mathbf{x})
    &= \operatorname{erf}(\alpha\mathbf{x}+s)
    \odot\boldsymbol{\gamma}
    +\boldsymbol{\beta} \\
    &\approx
    (\alpha\mathbf{x}+s)
    \odot
    \mathcal{R}(\alpha\mathbf{x}+s)
    \odot\boldsymbol{\gamma}
    +\boldsymbol{\beta} \\
    &=
    \left[
    \alpha\boldsymbol{\gamma}
    \odot
    \mathcal{R}(\alpha\mathbf{x}+s)
    \right]
    \odot\mathbf{x}
    +
    s\boldsymbol{\gamma}
    \odot
    \mathcal{R}(\alpha\mathbf{x}+s)
    +\boldsymbol{\beta} \\
    &= \mathbf{a}(\mathbf{x})\odot\mathbf{x}
    + \mathbf{b}(\mathbf{x}), \\[4pt]
    \mathbf{a}(\mathbf{x})
    &=
    \alpha\boldsymbol{\gamma}
    \odot
    \mathcal{R}(\alpha\mathbf{x}+s), \\
    \mathbf{b}(\mathbf{x})
    &=
    s\boldsymbol{\gamma}
    \odot
    \mathcal{R}(\alpha\mathbf{x}+s)
    +\boldsymbol{\beta}.
\end{aligned}
\label{derf}
\end{equation}

For $z=0$, $\mathcal{R}(z)$ is defined by continuity, with $\mathcal{R}(0)=4a_1$. And for a vector input, $\mathcal{R}(\cdot)$ is applied element-wise.

\subsubsection{SeeDNorm}

\begin{equation}
\begin{aligned}
    \operatorname{SeeDNorm}(\mathbf{x})
    &=
    \frac{\mathbf{x}}
    {\sqrt{\operatorname{Mean}(\mathbf{x}\odot\mathbf{x})+\epsilon}}
    \odot
    \left[
    \sigma\left(
    \mathbf{x}\boldsymbol{\beta}^{\mathrm{T}}
    \right)\boldsymbol{\alpha}
    +\boldsymbol{\gamma}
    \right] \\
    &=
    \frac{
    \sigma\left(
    \mathbf{x}\boldsymbol{\beta}^{\mathrm{T}}
    \right)\boldsymbol{\alpha}
    +\boldsymbol{\gamma}
    }
    {\sqrt{\operatorname{Mean}(\mathbf{x}\odot\mathbf{x})+\epsilon}}
    \odot\mathbf{x} \\
    &= \mathbf{a}(\mathbf{x})\odot\mathbf{x}
    + \mathbf{b}(\mathbf{x}), \\[4pt]
    \mathbf{a}(\mathbf{x})
    &=
    \frac{
    \sigma\left(
    \mathbf{x}\boldsymbol{\beta}^{\mathrm{T}}
    \right)\boldsymbol{\alpha}
    +\boldsymbol{\gamma}
    }
    {\sqrt{\operatorname{Mean}(\mathbf{x}\odot\mathbf{x})+\epsilon}}, \\
    \mathbf{b}(\mathbf{x})
    &= \mathbf{0}.
\end{aligned}
\label{seednorm}
\end{equation}

\subsection{Experiements about LENSNorm and LENSNorm+}
\label{lensnorm_plus_section}

\begin{table}[h]
    \centering
    \caption{Experimental Results on \emph{CIFAR-10} and \emph{CIFAR-100} using a Single-Head ViT with~\emph{LayerNorm},~\emph{Softmax},~\emph{LENSNorm},~\emph{LENSNorm+}
    and~\emph{LENSMax}.}
    \vspace{5pt}
    \label{lensnorm_plus_results}
    \resizebox{1\columnwidth}{!}
    {
        \begin{tabular}{cccccccc}
            \toprule
            Dataset                     & Training Path                             & Epoch               & Peak LR                           & Norm.    & Attn.   & Top-1 Acc. (\%) & $\Delta$Acc. (pp) \\  \midrule
            \multirow{4}{*}{CIFAR-10}  & \multirow{4}{*}{Scratch$\rightarrow$LENS} & \multirow{4}{*}{50} & \multirow{4}{*}{$5\times10^{-4}$} & LENSNorm  & Softmax & $81.41\pm0.17$  & 0.00 (Baseline)   \\
                                       &                                           &                     &                                   & LENSNorm+ & Softmax & $81.64\pm0.26$  & +0.23             \\
                                       &                                           &                     &                                   & LENSNorm  & LENSMax & $81.00\pm0.19$  & 0.00 (Baseline)   \\
                                       &                                           &                     &                                   & LENSNorm+ & LENSMax & $80.61\pm1.71$  & -0.39             \\ \midrule[1.5pt]
            \multirow{4}{*}{CIFAR-100} & \multirow{4}{*}{Scratch$\rightarrow$LENS} & \multirow{4}{*}{50} & \multirow{4}{*}{$5\times10^{-4}$} & LENSNorm  & Softmax & $52.14\pm0.83$  & 0.00 (Baseline)   \\
                                       &                                           &                     &                                   & LENSNorm+ & Softmax & $52.66\pm0.86$  & +0.52             \\
                                       &                                           &                     &                                   & LENSNorm  & LENSMax & $52.05\pm0.60$  & 0.00 (Baseline)   \\
                                       &                                           &                     &                                   & LENSNorm+ & LENSMax & $52.85\pm0.75$  & +0.80             \\ \bottomrule
        \end{tabular}
    }
    \vspace{5pt}
    \parbox{\linewidth}{\footnotesize \textbf{Note:} The experimental setting is the same as in \autoref{cifar_results}.}
    
\end{table}

\subsection{Conversion from Softmax to LENSMax}
\label{math_max}

Given an input vector $\mathbf{x}=[x_1,\ldots,x_N]$ as one row in $\mathbf{Q}\mathbf{K}^{\mathsf T}/\sqrt{d}$, the $i$-th output of \emph{Softmax} is
\begin{equation}
    \operatorname{Softmax}(\mathbf{x})_i
    =
    \frac{\exp(x_i)}
    {\sum_{j=1}^{N}\exp(x_j)}.
\end{equation}
We define two functions as
\begin{equation}
    \lambda(\mathbf{x})
    =
    \sum_{j=1}^{N}\exp(x_j),
    \qquad
    \mu(\mathbf{x})
    =
    \log\lambda(\mathbf{x}).
\end{equation}
The \emph{Softmax} output can then be reorganized as
\begin{equation}
\begin{aligned}
    \operatorname{Softmax}(\mathbf{x})_i
    &=
    \frac{\exp(x_i)}{\lambda(\mathbf{x})} \\
    &=
    \frac{\exp(x_i)}{\exp(\log\lambda(\mathbf{x}))} \\
    &=
    \exp\left(x_i-\mu(\mathbf{x})\right).
\end{aligned}
\end{equation}

\begin{equation}
    f(\mathbf{x})=\mu(\mathbf{x}),
    \qquad
    g(x_i,f(\mathbf{x}))
    =
    \exp\left(x_i-f(\mathbf{x})\right).
\end{equation}

\begin{equation}
    \operatorname{Softmax}(\mathbf{x})_i
    =
    g\left(x_i,f(\mathbf{x})\right).
\end{equation}

Inspired by the $g\left(x_i,f(\mathbf{x})\right)$ format of \emph{Softmax}, we define \emph{LENSMax} as one row-state function $r(\mathbf{x})$ and one element-mapping function $e(x_i,r(\mathbf{x}))$. 

To limit the output range of $e(x_i,r(\mathbf{x}))$, we apply a \emph{Sigmoid} function to its result and divide the \emph{Sigmoid} result by $2^{\left\lfloor\log_2N\right\rfloor}$, where $N$ denotes the number of valid elements in the row. Therefore, the sum of each \emph{LENSMax} output row is lower than 2:

\begin{equation}
    \sum_{i=1}^{N}\operatorname{LENSMax}(\mathbf{x})_i
    <
    \frac{N}{2^{\left\lfloor\log_2N\right\rfloor}}
    < 2.
\end{equation}

We then adopt two tiny MLPs to parameterize the row-state and element-mapping functions.~\emph{Sigmoid} is the activation function of the output layer in the MLP of the element-mapping function:

\begin{equation}
\begin{aligned}
    r(\mathbf{x})
    &=
    \operatorname{MLP}_1(\mathbf{x}),\\
    e(x_i,r(\mathbf{x}))
    &=
    \operatorname{MLP}_2\left(x_i,r(\mathbf{x})\right),\\
    \operatorname{LENSMax}(\mathbf{x})_i
    &=
    \frac{
    \operatorname{MLP}_2
    \left(x_i,\operatorname{MLP}_1(\mathbf{x})\right)}
    {2^{\left\lfloor\log_2N\right\rfloor}}.
\end{aligned}
\end{equation}

\subsection{Validation Accuracy Trajectory}
\label{training_plot}

\begin{figure}[h]
    \centering
    {
        \begin{minipage}{0.495\columnwidth}
            \centerline{\includegraphics[width=1\columnwidth]{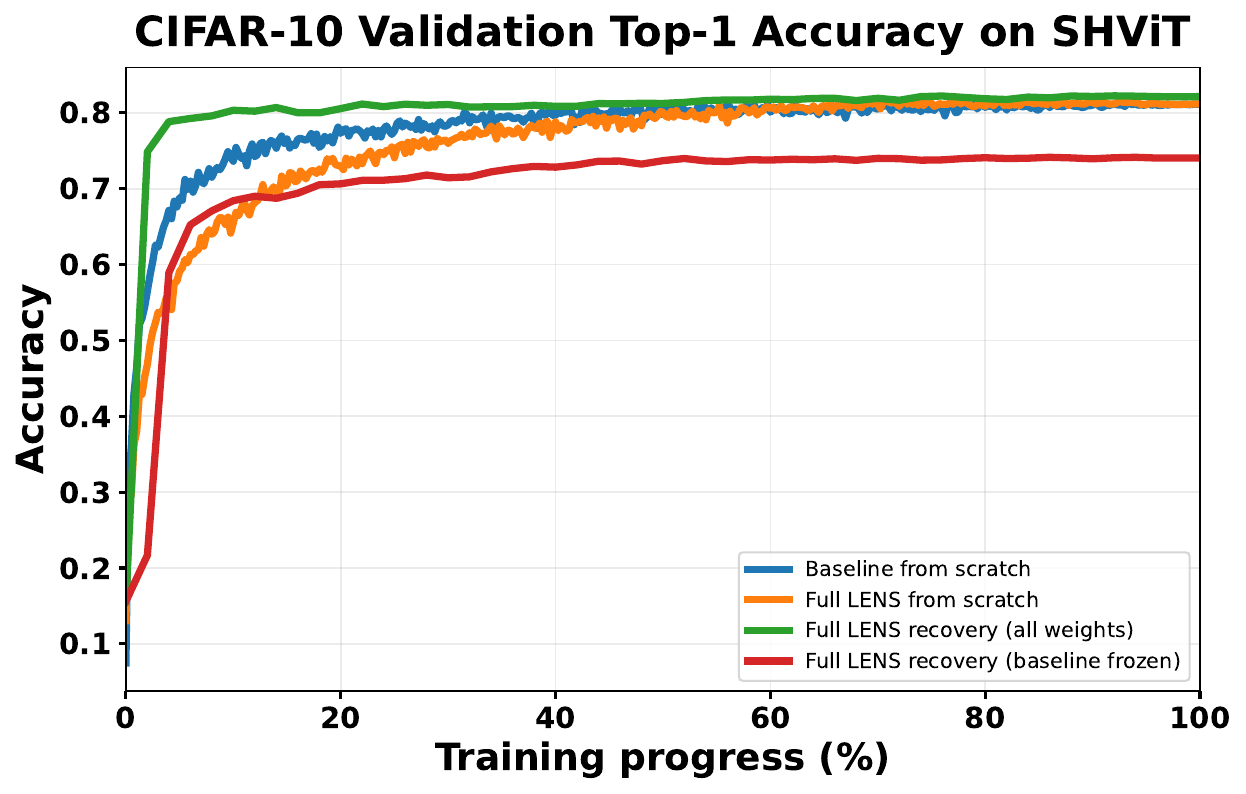}}
        \end{minipage}
        \begin{minipage}{0.495\columnwidth}
            \centerline{\includegraphics[width=1\columnwidth]{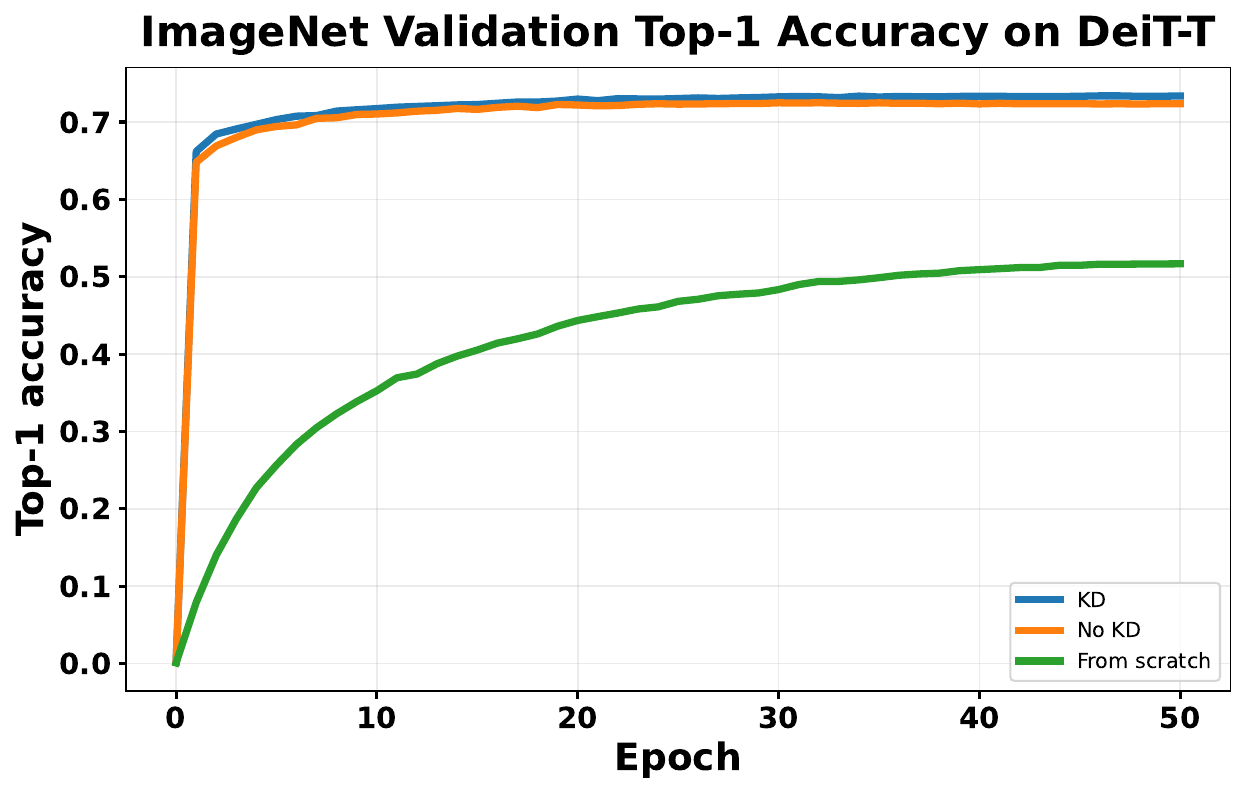}}
        \end{minipage}
    }
    \caption{Validation trajectory of training on single-head ViT and DeiT-Tiny.}
\end{figure}

\begin{figure}[t]
    \centering
    {
        \begin{minipage}{0.495\columnwidth}
            \centerline{\includegraphics[width=1\columnwidth]{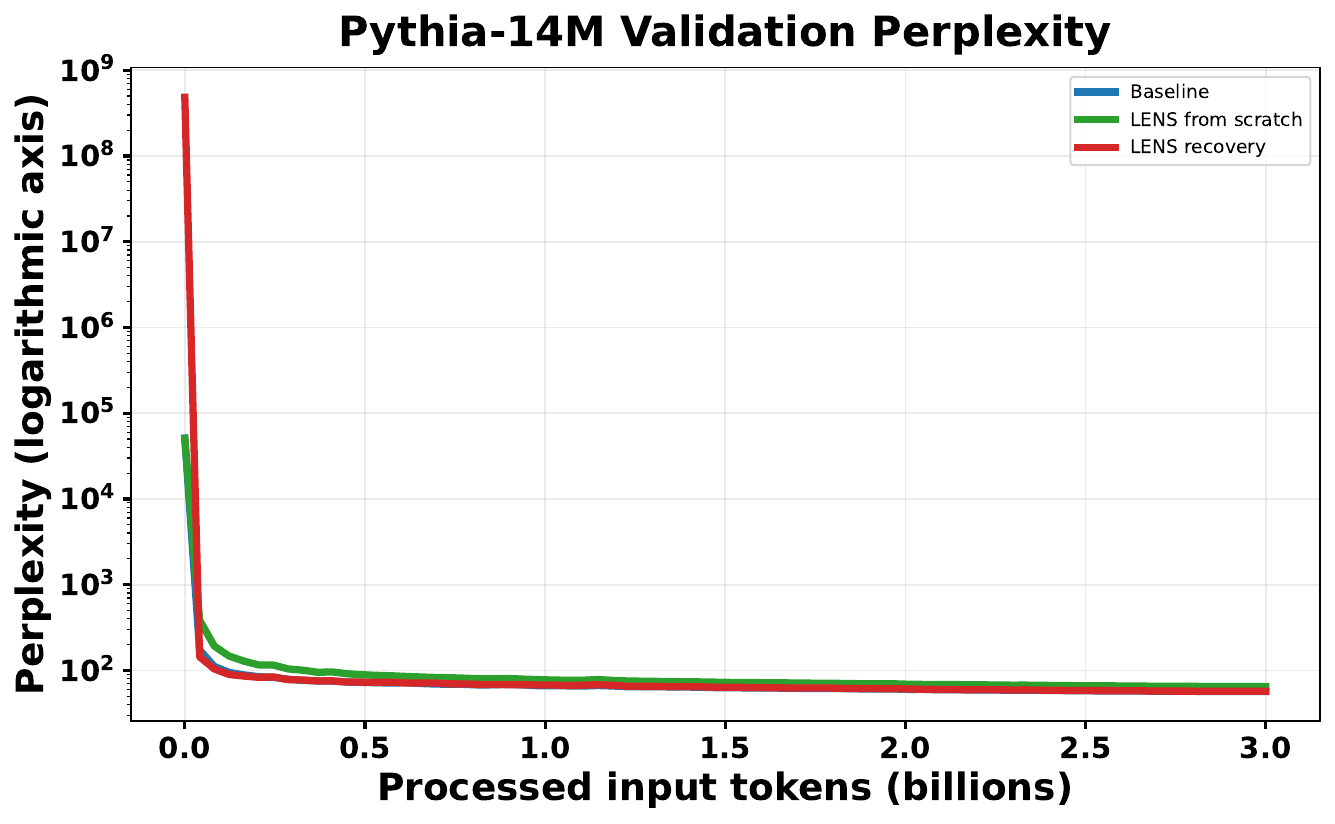}}
        \end{minipage}
        \begin{minipage}{0.495\columnwidth}
            \centerline{\includegraphics[width=1\columnwidth]{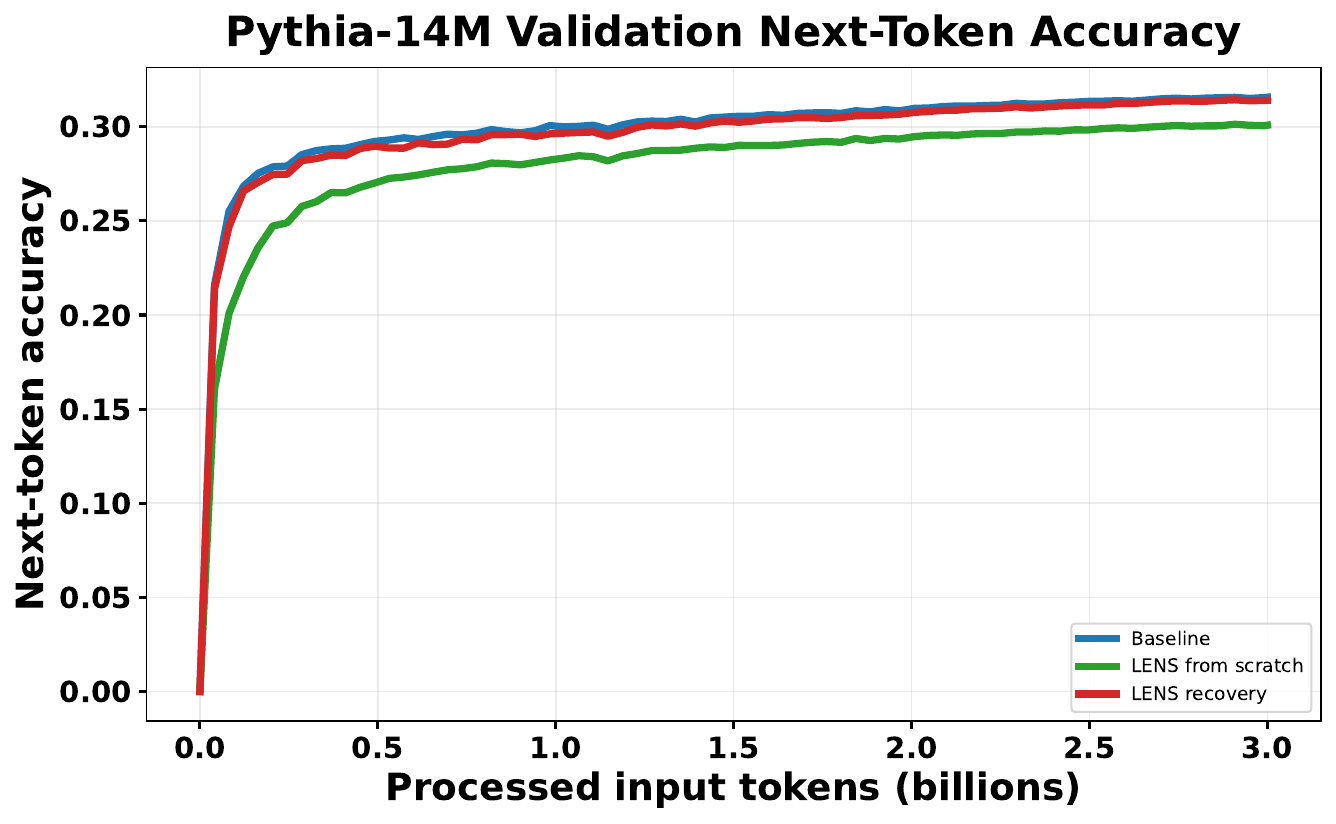}}
        \end{minipage}
    }
    \caption{Validation trajectory of training on Pythia-14M.}
    \label{val_in_training}
\end{figure}

We recorded the validation trajectory during training. The records show that the performance of all LENS models dropped to near-random levels immediately after replacement. The accuracy and language modeling performance of LENS models gradually recovered during retraining. This indicates that LENS replacement is not conventional fine-tuning, but joint retraining initialized from pretrained model parameters. Moreover, because replacing \emph{LayerNorm} and \emph{Softmax} with LENS causes a substantial architectural change and can produce extremely large gradients at the beginning of retraining, we clipped the global gradient norm to a maximum of $1.0$.

\subsection{Experiements about Replacing DyT/Derf/RMSNorm and Softmax with LENS}
\label{additional_result_section}

\begin{table}[h]
    \centering
    \caption{Experimental Results on \emph{CIFAR-10} and \emph{CIFAR-100} using a Single-Head ViT with~\emph{DyT/Derf/RMSNorm},~\emph{Softmax},~\emph{LENSNorm}
    and~\emph{LENSMax}.}
    \vspace{5pt}
    \label{additional_results}
    \resizebox{1\columnwidth}{!}
    {
        \begin{tabular}{cccccccc}
            \toprule
                Dataset                     & Training Path                             & Epoch               & Peak LR                           & Norm.    & Attn.   & Top-1 Acc. (\%) & $\Delta$Acc. (pp) \\  \midrule
                \multirow{12}{*}{CIFAR-10}  & From Scratch                              & 400                 & $1\times10^{-3}$                  & DyT      & Softmax & 76.75           & 0.00 (Baseline)   \\ \cmidrule(lr){2-8}
                                            & \multirow{3}{*}{Scratch$\rightarrow$LENS} & \multirow{3}{*}{50} & \multirow{3}{*}{$5\times10^{-4}$} & LENSNorm & Softmax & 76.97           & 0.22              \\
                                            &                                           &                     &                                   & DyT      & LENSMax & 76.66           & -0.09             \\
                                            &                                           &                     &                                   & LENSNorm & LENSMax & 78.15           & 1.4               \\ \cmidrule(lr){2-8}
                                            & From Scratch                              & 400                 & $1\times10^{-3}$                  & Derf     & Softmax & 77.00           & 0.00 (Baseline)   \\ \cmidrule(lr){2-8}
                                            & \multirow{3}{*}{Scratch$\rightarrow$LENS} & \multirow{3}{*}{50} & \multirow{3}{*}{$5\times10^{-4}$} & LENSNorm & Softmax & 78.61           & 1.61              \\
                                            &                                           &                     &                                   & Derf     & LENSMax & 76.20           & -0.80             \\
                                            &                                           &                     &                                   & LENSNorm & LENSMax & 78.92           & 1.92              \\ \cmidrule(lr){2-8}
                                            & From Scratch                              & 400                 & $1\times10^{-3}$                  & RMSNorm  & Softmax & 80.15           & 0.00 (Baseline)   \\ \cmidrule(lr){2-8}
                                            & \multirow{3}{*}{Scratch$\rightarrow$LENS} & \multirow{3}{*}{50} & \multirow{3}{*}{$5\times10^{-4}$} & LENSNorm & Softmax & 81.43           & 1.28              \\
                                            &                                           &                     &                                   & RMSNorm  & LENSMax & 80.33           & 0.18              \\
                                            &                                           &                     &                                   & LENSNorm & LENSMax & 80.84           & 0.69              \\ \midrule[1.5pt]
                \multirow{12}{*}{CIFAR-100} & From Scratch                              & 400                 & $1\times10^{-3}$                  & DyT      & Softmax & 45.84           & 0.00 (Baseline)   \\ \cmidrule(lr){2-8}
                                            & \multirow{3}{*}{Scratch$\rightarrow$LENS} & \multirow{3}{*}{50} & \multirow{3}{*}{$5\times10^{-4}$} & LENSNorm & Softmax & 49.39           & 3.55              \\
                                            &                                           &                     &                                   & DyT      & LENSMax & 46.27           & 0.43              \\
                                            &                                           &                     &                                   & LENSNorm & LENSMax & 49.66           & 3.82              \\ \cmidrule(lr){2-8}
                                            & From Scratch                              & 400                 & $1\times10^{-3}$                  & Derf     & Softmax & 46.24           & 0.00 (Baseline)   \\ \cmidrule(lr){2-8}
                                            & \multirow{3}{*}{Scratch$\rightarrow$LENS} & \multirow{3}{*}{50} & \multirow{3}{*}{$5\times10^{-4}$} & LENSNorm & Softmax & 49.94           & 3.70              \\
                                            &                                           &                     &                                   & Derf     & LENSMax & 46.57           & 0.33              \\
                                            &                                           &                     &                                   & LENSNorm & LENSMax & 49.09           & 2.85              \\ \cmidrule(lr){2-8}
                                            & From Scratch                              & 400                 & $1\times10^{-3}$                  & RMSNorm  & Softmax & 51.95           & 0.00 (Baseline)   \\ \cmidrule(lr){2-8}
                                            & \multirow{3}{*}{Scratch$\rightarrow$LENS} & \multirow{3}{*}{50} & \multirow{3}{*}{$5\times10^{-4}$} & LENSNorm & Softmax & 51.72           & -0.23             \\
                                            &                                           &                     &                                   & RMSNorm  & LENSMax & 51.99           & 0.04              \\
                                            &                                           &                     &                                   & LENSNorm & LENSMax & 52.43           & 0.48              \\ \bottomrule
        \end{tabular}
    }
    \vspace{5pt}
    \parbox{\linewidth}{\footnotesize \textbf{Note:} The experimental setting is the same as in \autoref{cifar_results}, except that the random seed is set to 42.}
    
\end{table}

\subsection{Relative L2 of Inherited Weights after LENS Recovery Training}

To quantify the changes in the inherited model parameters during LENS retraining, we compute the relative $\ell_2$ weight change as $|\mathbf{W}{\mathrm{LENS}}-\mathbf{W}{\mathrm{Base}}|2/|\mathbf{W}{\mathrm{Base}}|_2$ for all shape-compatible parameters shared by the baseline and retrained LENS models. 

For each shape-compatible parameter tensor, let $\mathbf{W}_{\mathrm{Base}}$ and $\mathbf{W}_{\mathrm{LENS}}$ denote its values in the baseline and retrained LENS models, respectively. We calculate its relative $\ell_2$ change as

\begin{equation}
    D_{\mathrm{rel}}(\mathbf{W})
    =
    \frac{
    \left\|
    \operatorname{vec}(\mathbf{W}_{\mathrm{LENS}})
    -
    \operatorname{vec}(\mathbf{W}_{\mathrm{Base}})
    \right\|_2
    }{
    \left\|
    \operatorname{vec}(\mathbf{W}_{\mathrm{Base}})
    \right\|_2+\epsilon
    },
\end{equation}

where $\operatorname{vec}(\cdot)$ flattens the parameter tensor and $\epsilon$ is a small constant for numerical stability. Equivalently, for a tensor containing $M$ elements,

\begin{equation}
    D_{\mathrm{rel}}(\mathbf{W})
    =
    \frac{
    \sqrt{\sum_{i=1}^{M}
    \left(
    W_{\mathrm{LENS},i}-W_{\mathrm{Base},i}
    \right)^2}
    }{
    \sqrt{\sum_{i=1}^{M}W_{\mathrm{Base},i}^{2}}
    +\epsilon
    }.
\end{equation}

A value of zero indicates no weight change, while a value of one means that the $\ell_2$ norm of the weight difference is equal to the $\ell_2$ norm of the original parameter tensor.

In our experiment, newly introduced LENS parameters and replaced \emph{LayerNorm} parameters are excluded. 

The results show substantial and structured changes rather than uniformly small updates. For example, the relative changes of the first-block QKV and attention-output projection reach 0.303 and 0.486 on CIFAR-10. Similar changes are observed in DeiT, including approximately 0.542 for the final QKV layer of DeiT-Tiny and approximately 1.0 for the final FFN layer of DeiT-Small. Together with the poor \textit{Only-LENS} results, these weight changes suggest that LENS jointly adapts with the inherited model parameters during retraining rather than merely approximating the replaced operators.

\begin{figure}[!h]
    \centering
    \begin{minipage}{0.9\columnwidth}
        \centerline{\includegraphics[width=1\columnwidth]{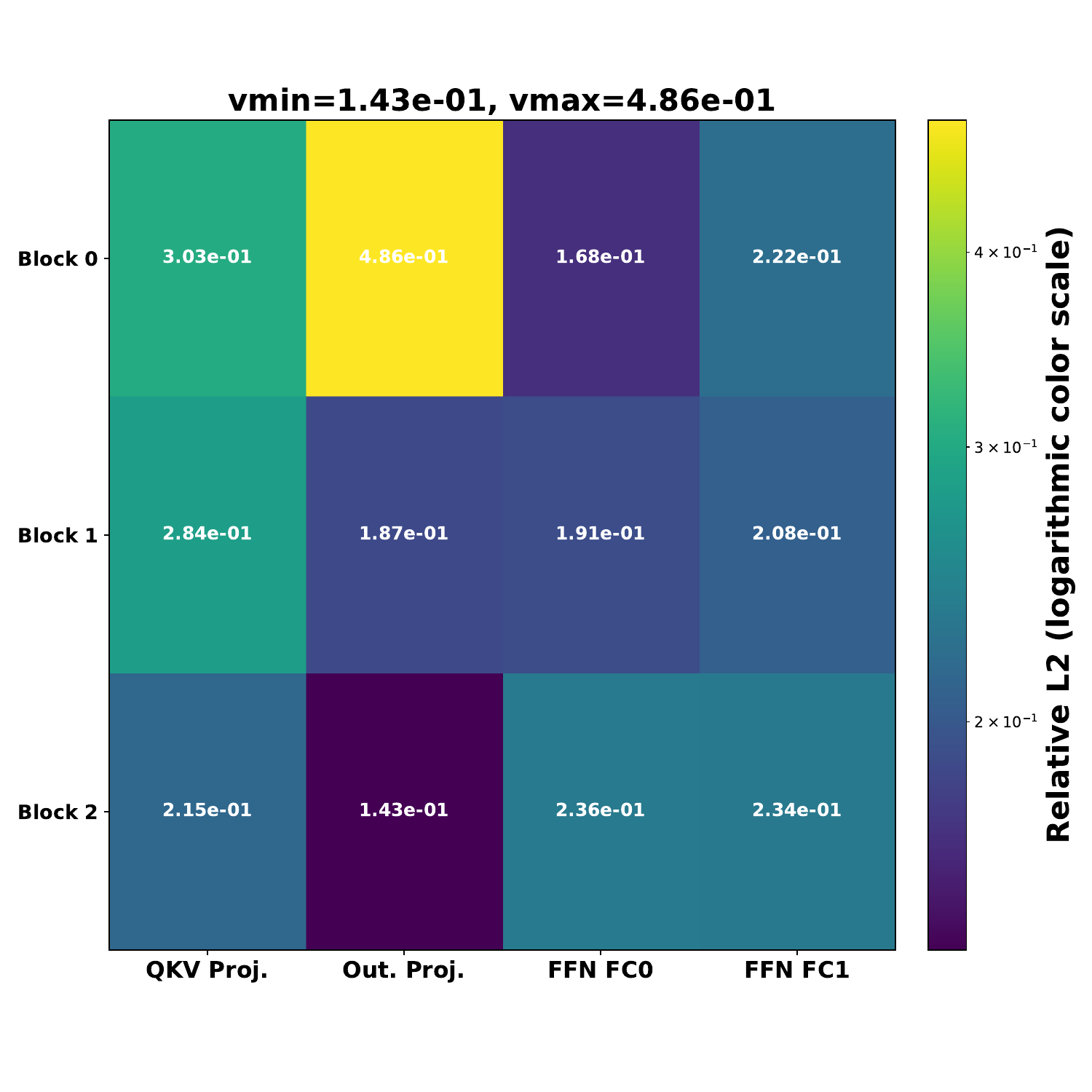}}
    \end{minipage}
    \label{relative_l2}
    \caption{Relative $\ell_2$ of inherited weights after LENS recovery training on single-head ViT using CIFAR-10.}
    \vspace{-20pt}
\end{figure}

\begin{figure}[!t]
    \centering
    \begin{minipage}{0.9\columnwidth}
        \centerline{\includegraphics[width=1\columnwidth]{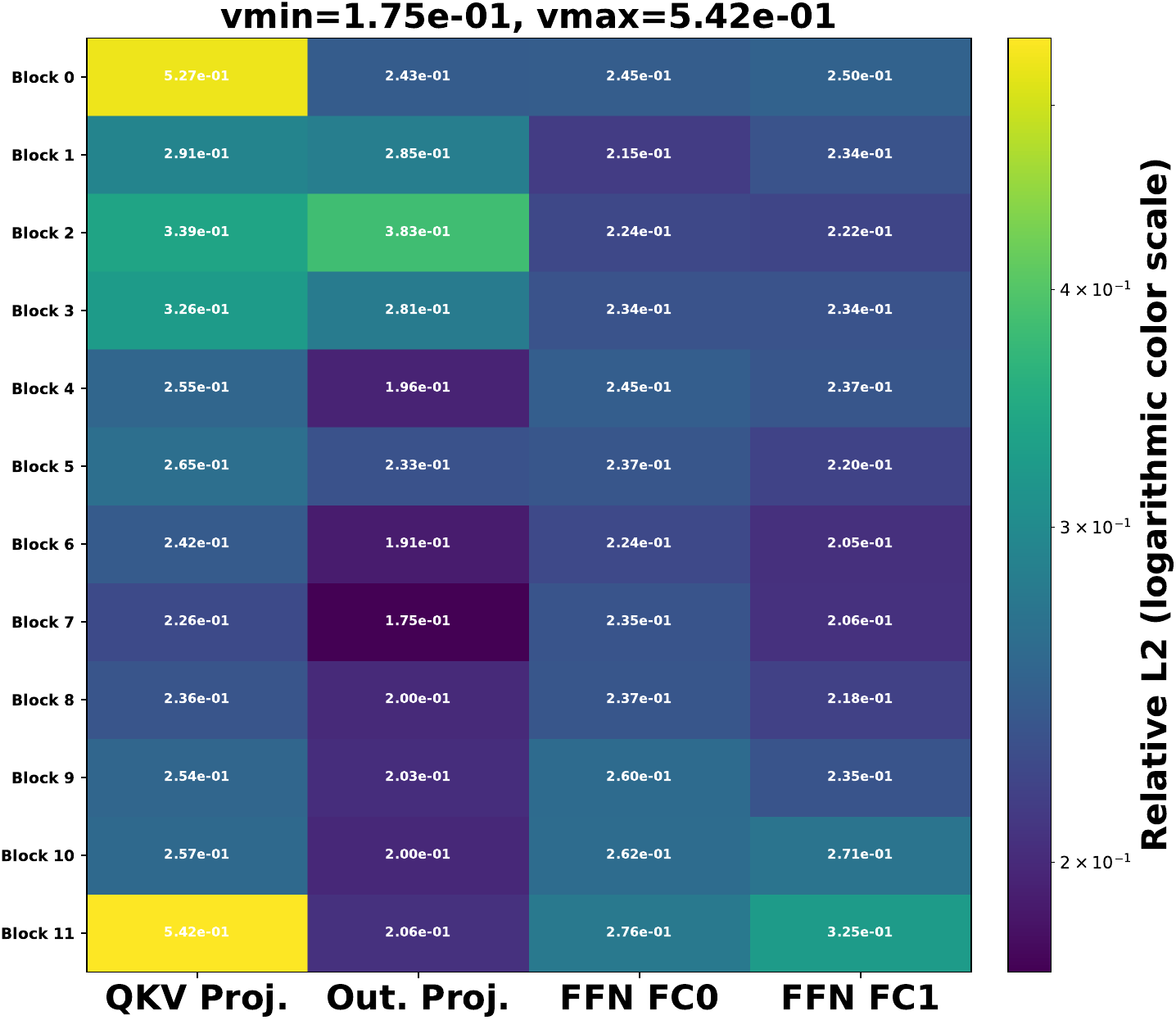}}
    \end{minipage}
    \label{relative_l2}
    \caption{Relative $\ell_2$ of inherited weights after LENS recovery training on DeiT-Tiny.}

\end{figure}

\begin{figure}[!t]
    \centering
    \begin{minipage}{0.9\columnwidth}
        \centerline{\includegraphics[width=1\columnwidth]{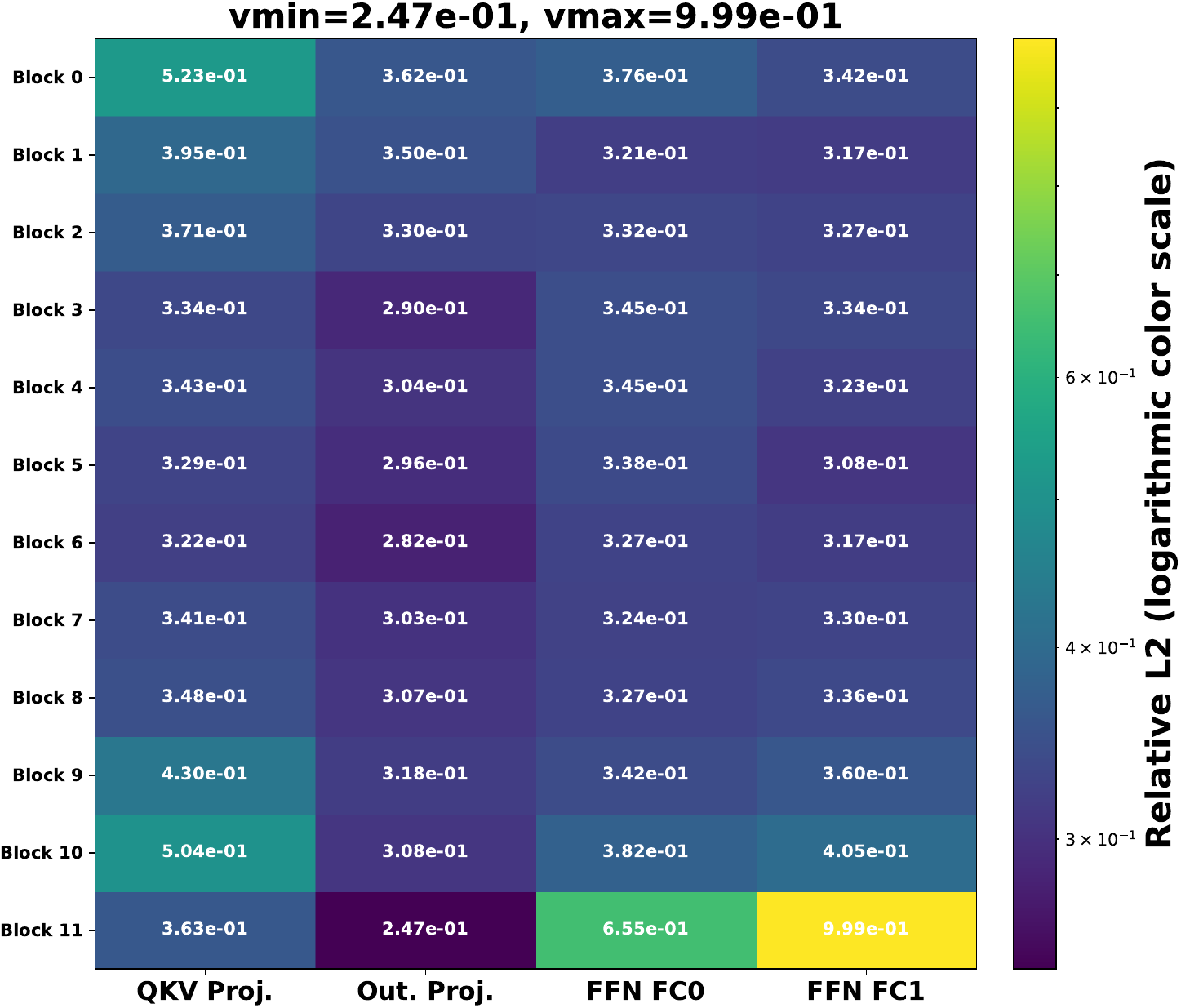}}
    \end{minipage}
    \label{relative_l2}
    \caption{Relative $\ell_2$ of inherited weights after LENS recovery training on DeiT-Small.}
    \vspace{-20pt}
\end{figure}

\begin{figure}[!t]
    \centering
    \begin{minipage}{0.9\columnwidth}
        \centerline{\includegraphics[width=1\columnwidth]{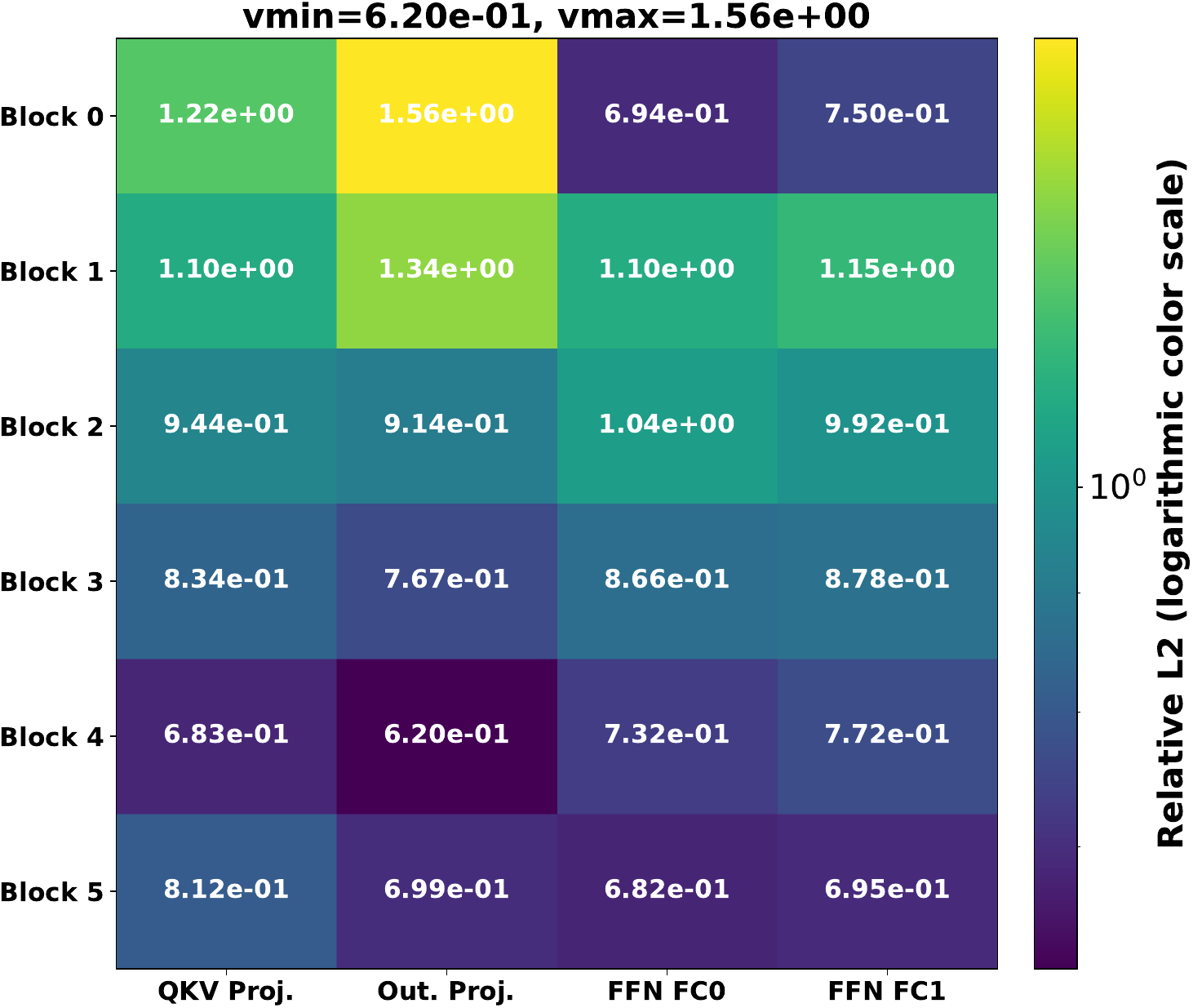}}
    \end{minipage}
    \label{relative_l2}
    \caption{Relative $\ell_2$ of inherited weights after LENS recovery training on Pythia-70M.}
\end{figure}

\end{document}